\documentclass[lettersize,journal]{IEEEtran}
\usepackage{amsmath,amsfonts,amssymb}
\usepackage[ruled,linesnumbered]{algorithm2e}
\usepackage{algorithmic}
\usepackage{array}
\usepackage[caption=false,font=normalsize,labelfont=sf,textfont=sf]{subfig}
\usepackage{textcomp}
\usepackage{stfloats}
\usepackage{url}
\usepackage{verbatim}
\usepackage{graphicx}
\usepackage{booktabs}
\usepackage{cite}
\usepackage[table]{xcolor}
\usepackage{hyperref}
\usepackage{cleveref}

\usepackage{multirow}
\usepackage{makecell}
\usepackage{colortbl}
\usepackage{xspace}

\newcommand{\ie}{\textit{i.e.}\@\xspace}

\newcommand{\la}[1]{{\color{black}#1}} 
\newcommand{\hl}[1]{{\color{black}#1}}

\begin{document}

\title{Progressive Human Motion Generation \\ Based on Text and Few Motion Frames}

\author{Ling-An Zeng, Gaojie Wu, Ancong Wu, Jian-Fang Hu, Wei-Shi Zheng
\IEEEcompsocitemizethanks{\IEEEcompsocthanksitem
Ling-An Zeng is with the School of Artificial Intelligence, Sun Yat-sen University, Zhuhai, Guangdong 519082, China (e-mail: zenglan3@mail2.sysu.edu.cn). 
Gaojie Wu, Ancong Wu, and Jian-Fang Hu are with the School of Computer Science and Engineering, Sun Yat-sen University, Guangzhou, Guangdong 510275, China (e-mail: wugj7@mail2.sysu.edu.cn; wuanc@mail.sysu.edu.cn; hujf5@mail.sysu.edu.cn).
\IEEEcompsocthanksitem
Wei-Shi Zheng is with the School of Computer Science and Engineering, Sun Yat-sen University, China; the Guangdong Key Laboratory of Information Security Technology, China; the Key Laboratory of Machine Intelligence and Advanced Computing, Ministry of Education, China; and Peng Cheng Laboratory, China (e-mail: wszheng@ieee.org; zhwshi@mail.sysu.edu.cn).
\IEEEcompsocthanksitem Ancong Wu is the corresponding author.
\IEEEcompsocthanksitem Copyright © 2025 IEEE. Personal use of this material is permitted. However, permission to use this material for any other purposes must be obtained from the IEEE by sending an email to pubs-permissions@ieee.org.
}}



\maketitle

\begin{abstract}
Although existing text-to-motion (T2M) methods can produce realistic human motion from text description, it is still difficult to align the generated motion with the desired postures since using text alone is insufficient for precisely describing diverse postures. To achieve more controllable generation, an intuitive way is to allow the user to input a few motion frames describing precise desired postures. Thus, we explore a new Text-Frame-to-Motion (TF2M) generation task that aims to generate motions from text and very few given frames. Intuitively, the closer a frame is to a given frame, the lower the uncertainty of this frame is when conditioned on this given frame. Hence, we propose a novel Progressive Motion Generation (PMG) method to progressively generate a motion from the frames with low uncertainty to those with high uncertainty in multiple stages. During each stage, new frames are generated by a Text-Frame Guided Generator conditioned on frame-aware semantics of the text, given frames, and frames generated in previous stages. Additionally, to alleviate the train-test gap caused by multi-stage accumulation of incorrectly generated frames during testing, we propose a Pseudo-frame Replacement Strategy for training. Experimental results show that our PMG outperforms existing T2M generation methods by a large margin with even one given frame, validating the effectiveness of our PMG. Code is available \href{https://github.com/qinghuannn/PMG}{here}.
\end{abstract}


\begin{figure}[t]
    \centering
    \includegraphics[width=\linewidth]{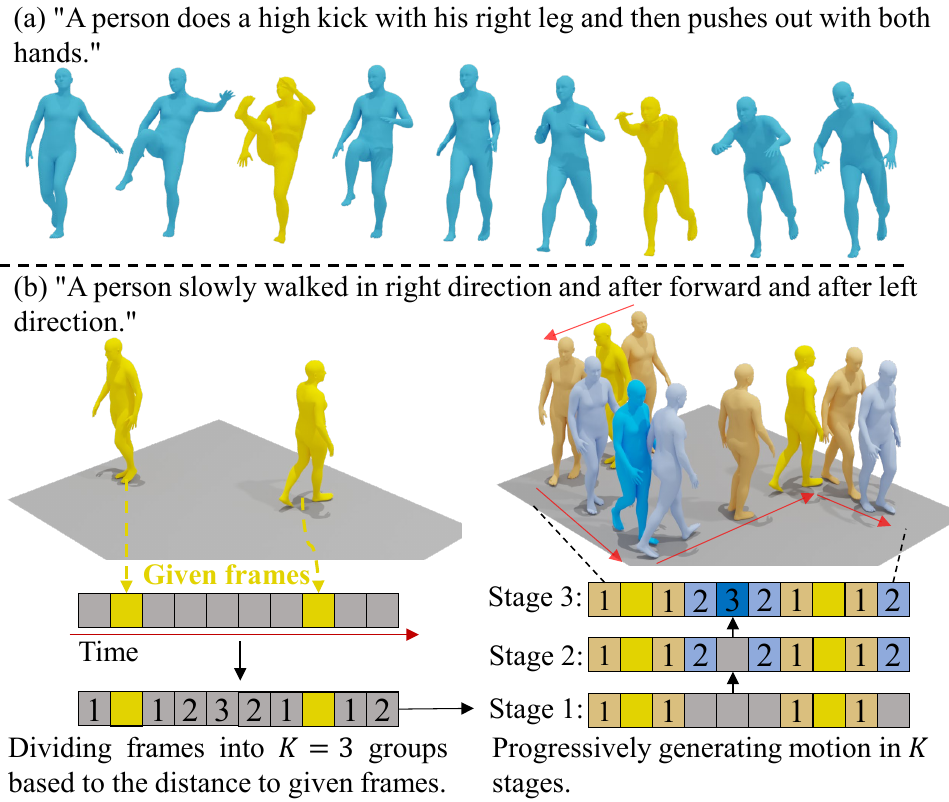}
    \caption{\hl{Illustration of the Text-Frame-to-Motion generation task (a) and our proposed method (b).  This task focuses on generating motions from text descriptions and very few given frames. The given frames are in yellow.} }
    \label{fig:intro}
\end{figure}

\section{Introduction}

\IEEEPARstart{R}{ecent} text-to-motion (T2M) generation methods \cite{t2m, zhang2023remodiffuse, momask, bamm,chainhoi,tang2023temporal,sun2021action,zhang2024modular,zhang2024generative,zeng2020hybrid,zeng2024multimodal} have achieved significant progress and can generate realistic 3D human motion from text descriptions, showing significant implications for AR/VR, gaming,  and film production \cite{nazarieh2024survey,10634078}. Nevertheless, accurately aligning the generated motion with desired postures remains challenging, as relying solely on text descriptions proves inadequate for precisely describing diverse postures. For example, the human postures corresponding to ``\textit{high kick}'' are infinitely varied, so we cannot generate a motion with the desired human postures just from the word ``\textit{high kick}''.

\hl{To achieve more controllable generation, an intuitive way is to allow the user to input a few motion frames describing desired postures. As shown in \Cref{fig:intro}.a, the text controls the motion at a high level, and the given frames spatially control the desired human postures. Thus, in this work, we propose a more controllable motion generation task, called the Text-Frame-to-Motion (TF2M) generation task, aiming to generate human motions from text and a few motion frames.} We assume that the given frames are relevant to the text since they are used to specify desired postures that are difficult to describe with text.

Different from T2M, the generated motion on TF2M should not only semantically match the given text but also spatially match the given frames, indicating the core of TF2M is how to utilize given frames effectively. Inspired by keyframe animation, an intuitive method is to generate the motion frames progressively based on the distance to the nearest given frame. It is obvious that the closer a frame is to a given frame, the lower the uncertainty of this frame is when conditioned on this given frame. However, different from keyframe animation requiring lots of keyframes, TF2M focuses on generating motion from extremely few frames since obtaining 3D human postures is not easy. Nevertheless, limited given frames still provide certain information about the desired poses.

Based on the above consideration, we propose a novel generation paradigm named \textbf{Progressive Motion Generation} to progressively generate a motion from the frames with low uncertainty to those with high uncertainty in multiple stages. As shown in \Cref{fig:intro}.b, our method progressively generates the target motion in multiple stages and divides motion frames to be generated into several groups corresponding to different stages based on their distance to the nearest given frame. During the $k$-th stage, the frames in the $k$-th group are generated conditioned on the text description and obtained frames, including given frames and frames generated in previous stages.

To generate frames in the $k$-th group, we propose a novel \textbf{Text-Frame Guided Generator} for conditional frame generation. Specifically, we propose a Frame-aware Semantics Decoder to extract frame-aware semantics, since only a part of the semantics of the text is relevant to the frames being generated, while the obtained frames indicate what has already been generated. Additionally, we design a Text-Frame Guided Block to generate new frames conditioned on frame-aware semantics and obtained frames, which explores the relationship between new frames and obtained frames and then uses frame-aware semantics to guide generation.

In order to address the train-test gap arising from the multi-stage accumulation of incorrectly generated frames during testing, we propose a \textbf{Pseudo-frame Replacement Strategy} for training. Specifically, we randomly replace some motion frames with those generated by a momentum-based moving average of our generator during training. In this way, our method is able to generate more accurate frames conditioned on not entirely accurate frames generated in previous stages.

Formally, we call our method \textbf{Progressive Motion Generation} (PMG), which generates motions via the Progressive Motion Generation paradigm and a Text-Frame Guided Generator. Extensive experiments on two public datasets, \ie, HumanML3D\cite{t2m} and KIT-ML\cite{kit}, show that our method outperforms existing T2M generation methods by a large margin (especially in FID) even with one given frame, demonstrating the effectiveness of our method. We also analyze the influence of different given frames, and the results show our method is robust when given frames are relevant to the text.

\hl{
In summary, our main contributions are as follows:
\begin{enumerate}
    \item Our work is the first focusing on more controllable motion generation, \ie, the Text-Frame-to-Motion (TF2M) generation task, which aims to utilize text and a few frames to guide human motion generation.
    \item We propose a novel Progressive Motion Generation method, consisting of a new generation paradigm and a Text-Frame Guided Generator, that generates frames progressively based on the distance to the nearest given frame.
    \item We design a Pseudo-frame Replacement Strategy for training to alleviate the train-test gap arising from the multi-stage accumulation of incorrectly generated frames during testing.
\end{enumerate}
}

\section{Related Work}
\label{relat}
\noindent\textbf{Text-to-motion Generation.}
\hl{The Text-to-Motion (T2M) generation task aims to generate 3D human motion driven by text input. Some works \cite{t2m, temos, attt2m, bridgegap} project motion into a latent space and then learn the relationship between text and motion within this latent space. Other approaches \cite{t2m-gpt, jiang2023motiongpt, humantomato, avatargpt, wandr} adopt VQ-VAE \cite{vqvae} to learn discrete representations, generating motion in an auto-regressive manner via a generative pre-trained transformer. Additionally, some methods use diffusion models to generate motion either in the raw space \cite{mdm, md, light-t2m, mofusion, flowMDM, emdm} or in the latent space \cite{mld, sampieri2024length, yu2020multimodal, li2024toward}. MoMask \cite{momask} and BAMM \cite{bamm} introduce generative masked modeling frameworks. Other works explore retrieval-augmented generation \cite{zhang2023remodiffuse}, body-part-aware generation \cite{parco}, and 2D image-enhanced generation \cite{CrossDiff}. However, due to the highly abstract nature of text, using text alone as a condition can make it challenging to accurately control the generated results \cite{nazarieh2024survey}. Different from these works, our work focuses on a more controllable motion generation.
}

\vspace{0.1cm}
\hl{
\noindent\textbf{Controllable Text-to-motion Generation.}
To achieve more controllable motion generation, some works explore the use of additional conditions to achieve more controllable motion generation.} For example, GMD \cite{karunratanakul2023gmd} utilizes keyframe locations and obstacle avoidance to control the generated motion. PriorMDM \cite{priorMDM}, OmniControl \cite{omnicontrol}, TLcontrol \cite{tlcontrol}, and MotionLCM \cite{motionlcm} can incorporate different key joint positions to control the generated motion. However, the conditional signals, including trajectory, obstacle avoidance, and key joint positions used in existing works are insufficient to fully describe the desired postures. Unlike existing methods, this work focuses on a more controllable motion generation task that aims to utilize text and extremely few given frames to generate human motion with desired postures. For this purpose, we propose a Progressive Motion Generation paradigm and a Text-Frame Guided Generator to fully use the given postures.

\vspace{0.1cm}
\noindent\textbf{Keyframe-to-motion Generation.}
\hl{
The keyframe-to-motion (K2M) generation task aims to generate motion based on given keyframes. Some works \cite{harvey2020robust, duan2021single} use sequence-to-sequence models to address the motion completion problem, which involves filling in frames between keyframes. Other approaches \cite{li2022ganimator, wen2018generating} leverage generative adversarial networks (GANs) to generate high-quality results. Additionally, RSMT \cite{tang2023rsmt} proposes an online framework for styled real-time in-between motion generation. However, the K2M task typically requires a large number of keyframes and is limited to generating short motions between only two keyframes. In contrast, TF2M aims to generate long and complex motions with an extremely limited number of given frames. Furthermore, in TF2M, motion generation is globally constrained by the text input, rather than solely by the few given frames.
}

\section{Progressive Motion Generation}

\subsection{Preliminaries}
\label{sec:pre}

\noindent\textbf{Problem Formulation.}
Given a text ${w_i}$ describing a motion and a few motion frames ${f_i}$ relevant to the text, the goal is to generate a motion sequence of 3D human poses $M={m_i}$, which semantically matches the text and spatially matches the given frames. The time positions $P^0={p_i}$ of the given frames are also needed. Following \hl{recent works \cite{momask, zhang2023remodiffuse, light-t2m}}, the target motion length $N$ is also required for generation. 
Since the motion representation proposed in \cite{t2m} utilizes linear or angular velocities to represent root joint information, it is insufficient to obtain the root information of the current frame from the information of a single frame. To address this problem, we represent the human posture as the concatenation of the motion representation proposed in \cite{t2m} and the absolute position information of the root joint, which includes the root angle along the Y-axis ($r^a\in \mathbb{R}$) and root position ($r^p\in \mathbb{R}^3$). For each given motion frame, only the absolute position information of the root joint and the local joint positions are provided.

\vspace{0.1cm}
\noindent\textbf{Diffusion Model.} \label{sec:diffusion}
Thanks to the success of diffusion models \cite{sohl2015deep, ho2020denoising}, we adopt diffusion models to generate motions. Diffusion models define a Markov chain of diffusion steps that gradually add random noise to data. They are then trained to reverse the diffusion process to construct samples from the noise.

Formally, given a data $\mathbf{x}_0 \sim q(\mathbf{x}_0)$, small Gaussian noises are added to the data in $T$ steps, producing a sequence of noisy data $\mathbf{x}_1, ..., \mathbf{x}_T$. Then, the \textit{forward process} can be defined as:
{
\begin{gather}
    q(\mathbf{x}_t|\mathbf{x}_{t-1}) = \mathcal{N}(\mathbf{x}_t; \sqrt{1-\beta_t}\mathbf{x}_{t-1}, \beta_t \mathbf{I}), \\
    q(\mathbf{x}_{1:T}|\mathbf{x}_0) = \prod_{t=1}^{T} q(\mathbf{x}_t|\mathbf{x}_{t-1}), \label{eq:forward}
\end{gather}
}\noindent
where the variance schedule $\{\beta_t\in(0, 1)\}_{t=1}^T$ controls the step sizes. 
The \textit{backward process} is defined as:
{
\begin{gather}
    p_\theta(\mathbf{x}_{t-1}|\mathbf{x}_t) = \mathcal{N}(\mathbf{x}_{t-1};\mu_{\theta}(\mathbf{x}_t, t), \sigma_{\theta}(\mathbf{x}_t, t)), \label{eq:qsample}\\
    p_\theta(\mathbf{x}_{0:T}) = p(\mathbf{x}_T)\prod_{t=1}^{T}p_\theta(\mathbf{x}_{t-1}|\mathbf{x}_t),
 \end{gather}
}\noindent
where $\theta$ denotes model parameters, and the mean $\mu_{\theta}(\mathbf{x}_t, t)$ and variance $ \sigma_{\theta}(\mathbf{x}_t, t)$ are parameterized by neural networks. We can sample data from a Gaussian noise $\mathbf{x}_T \sim \mathcal{N}(\textbf{0},\textbf{I})$ via the \textit{backward process}. Please refer to \cite{sohl2015deep, ho2020denoising} for more details.

\begin{figure*}[t]
    \centering
    \includegraphics[width=\linewidth]{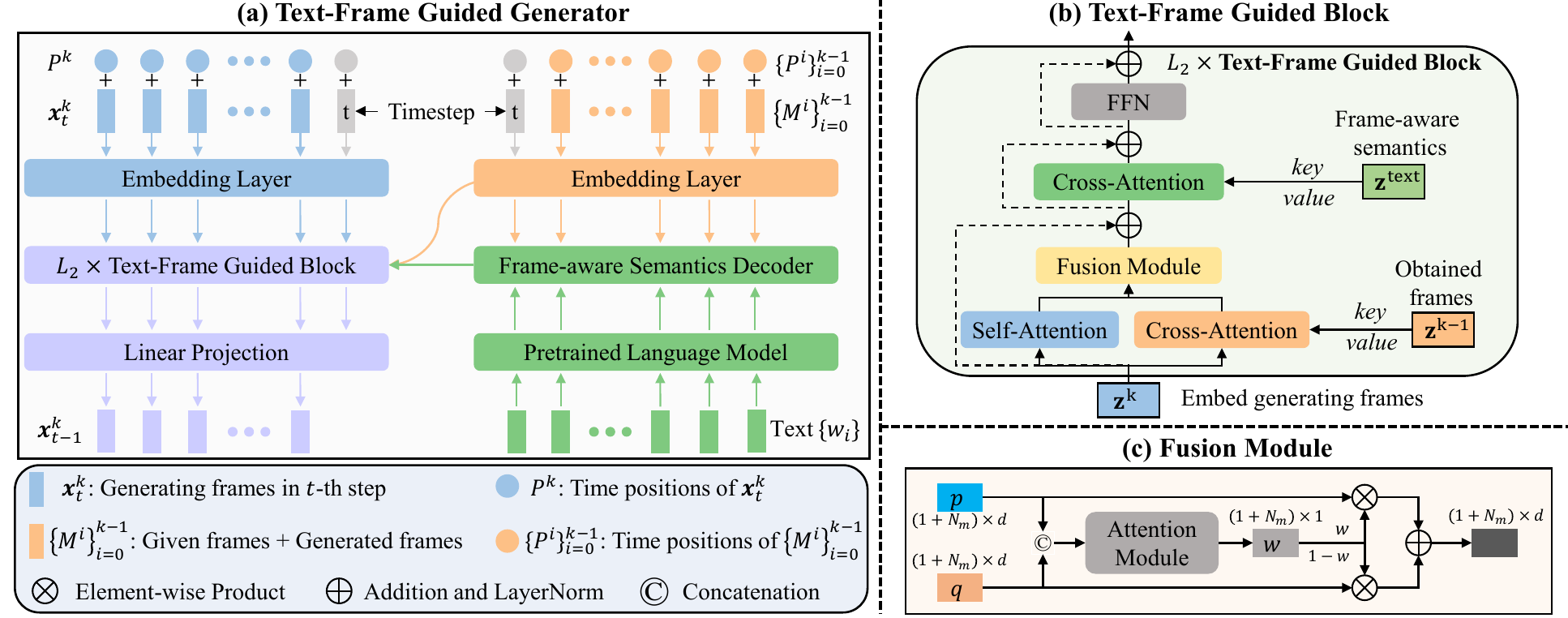}
    \caption{
    \hl{Overview of our Progressive Motion Generation (PMG). (a) Our PMG is a diffusion-based model and generates the target motion in $K$ stages, as shown in \Cref{fig:intro}. During the $k$-th stage, frames $M^k$ are generated via the denoising process, where $\mathbf{x}^k_T \sim \mathcal{N}(\textbf{0},\textbf{I})$. To generate $\mathbf{x}^k_{t-1}$ conditioned on the text, given frames, and frames generated in previous stages, we propose a Text-Frame Guided Generator, which mainly contains a Frame-aware Semantics Decoder and $L_2$ Text-Frame Guided Blocks. Note that $M^0$ denotes the given frames, and the pretrained language model is kept fixed.
    (b) Illustration of Text-Frame Guided Block. 
    (c) Illustration of Fusion Module. $N_m$ denotes the number of generating frames.}
    }
    \label{fig:overview}
\end{figure*}

\subsection{Progressive Motion Generation Paradigm}
\label{sec:pmgp}
To effectively utilize given motion frames, we design a novel Progressive Motion Generation paradigm to generate the motion frames progressively based on the distance to the nearest given frame in $K$ stages. 

\hl{Formally, suppose generating a motion $\{m_i\}^{N}_{i=1}$ in $K$ stages, we first uniformly divide the positions of target frames into $K$ groups based on the distance to the nearest given frame:}
{\abovedisplayskip=2pt \belowdisplayskip=2pt
\begin{gather}
    dis(i,P^0) = \min(\{|i - p_j| \; \big| \; p_j \in P^0\}), \label{eq:divide1} \\ 
    P^k = \{i \; \big| \; \left\lceil \frac{dis(i,P^0)}{dis(N,P^0)} * K \right\rceil = k, 1 \leq i \leq N \}, \label{eq:divide}
\end{gather} }\noindent
where $dis(N,P^0) = \max (\{dis(i,P^0)\}_{i=1}^{N})$, $P^0=\{p_i\}$ is the time positions of given frames, and $P^k$ is the time positions of frames in the $k$-th group. 

\hl{Then, we progressively generate the whole motion in $K$ stages. Specifically, in the $k$-th stage, $M^k$ can be generated as:}
\begin{equation}
    M^k = g_\theta(\{w_i\},  \{M^i\}_{i=0}^{k-1}, \{P^i\}_{i=0}^{k}),
\end{equation}
where $M^0 = \{f_i\}$ denotes given frames, $\{w_i\}$ is the text, and $g_\theta(\cdot)$ is our generation function. In this way, the entire motion $M$ is generated as:
{\abovedisplayskip=2pt \belowdisplayskip=2pt
\begin{equation}
    M = \bigcup_{k=0}^K M^k.
\end{equation}
}

\hl{In detail, to generate $M^k$ via the diffusion model,} the concatenation of frames in $M^k$ is treated as the target $\mathbf{x}_0^k \in \mathcal{R}^{|M^k| \times d_m}$. $\mathbf{x}^k_{t-1}$ can be got by:
\begin{equation}
    \mathbf{x}_{t-1}^k = g_\theta^t(\mathbf{x}_t^k, \{w_i\}, \{M^i\}_{i=0}^{k-1}, \{P^i\}_{i=0}^{k}). \label{eq:net}
\end{equation}

\hl{Furthermore, to achieve the conditional generation, we adopt the \textit{classifier-free guidance} method \cite{ho2022classifier}.} Specifically, $\mathbf{x}_{t-1}^k$ can be obtained via \Cref{eq:qsample} of the \textit{backward process}, where $\mu_{\theta}(\mathbf{x}_t^k, t)$ and variance $ \sigma_{\theta}(\mathbf{x}_t^k, t)$ are calculated by predicting a conditional noise $\hat{\epsilon}_\theta(\mathbf{x}_t^k, t, \bar{\mathbf{c}}^k)$.
\begin{equation}
    \hat{\epsilon}_\theta(\mathbf{x}_t^k, t, \bar{\mathbf{c}}^k) = (1+s) \; \epsilon_\theta(\mathbf{x}_t^k, t, \bar{\mathbf{c}}^k) - s \; \epsilon_\theta(\mathbf{x}_t^k, t, \hat{\mathbf{c}}^k), \label{eq:cfg}
\end{equation}
where $\hat{\mathbf{c}}^k = \{M^i\}_{i=0}^{k-1} \cup \{P^i\}_{i=0}^{k}$, $\bar{\mathbf{c}}^k = \hat{\mathbf{c}}^k \cup \{w_i\}$, and $s$ is a hyper-parameter that controls the strength of the conditions.

Actually, both $\epsilon_\theta(\mathbf{x}_t, t, \bar{\mathbf{c}}^k)$ and $\epsilon_\theta(\mathbf{x}_t, t, \hat{\mathbf{c}}^k)$ can be achieved via a single network, which is trained via the following \textit{training objective}:
\begin{equation}
    \mathcal{L} = \mathbb{E}_{\mathbf{x}^k_0, \epsilon}[|| \epsilon_t - \epsilon_{\theta}(\mathbf{x}_t^k, t, \mathbf{c}^k) ||^2], \label{eq:cfg2}
\end{equation}
where $\mathbf{c}^k$ denotes $\bar{\mathbf{c}}^k$ or $\hat{\mathbf{c}}^k$. During training, $\hat{\mathbf{c}}^k$ can be obtained by setting $\{w_i\}=\varnothing$. A hyper-parameter $\eta$ is used to control the text replacement probability. 

\vspace{0.1cm}
\noindent\textbf{Discussion.}
During inference, our method can generate motion under any number of given frames and any number of generation stages $K$. The number $K$ influences the inference times and the qualities of the generated motion, and we conduct experiments with different $K$ in \Cref{sec:ablation}.

\subsection{Text-Frame Guided Generator}\label{sec:arch}
To generate $\mathbf{x}_{t-1}^k$ conditioned on the text and obtained frames generated in previous stages via \Cref{eq:net}, we propose a Text-Frame Guided Generator mainly containing a Frames-aware Semantics Decoder (FSD) and several Text-Frame Guided Blocks (TGBs). \hl{As shown in \Cref{fig:overview}.a}, both obtained frames $\{M^i\}_{i=0}^{k-1}$ and generating frames $\mathbf{x}_{t-1}^k$ are fed into an embedding layer. The FSD generates frame-aware semantics by taking embedded obtained frames and the entire semantics of the text as inputs. Then, the TGBs generate $\mathbf{x}_{t-1}^k$ in embedding space conditioned on embedded obtained frames and frame-aware semantics. Finally, the outputs are projected into the actual motion representation.

\hl{Specifically, suppose that embedded obtained frames $\{M^i\}_{i=0}^{k-1}$ and embedded generating frames $\mathbf{x}_{t-1}^k$ are $\mathbf{z}^{k-1}$ and $\mathbf{z}^{k}$, respectively.} $\mathbf{z}^{k-1}$ and $\mathbf{z}^{k}$ can be got via:
{
\begin{equation}
    \mathbf{z} = proj(frames) + pos(positions),
\end{equation}
}\noindent
where $frames$ are $\{M^i\}_{i=0}^{k-1}$ or $\mathbf{x}_{t-1}^k$, $positions$ are $\{P^i\}_{i=0}^{k-1}$ or $P^k$, $proj(\cdot)$ is a linear projection, and $pos(\cdot)$ is a position embedding.

\vspace{0.15cm}
\noindent\textbf{Frames-aware Semantics Decoder.}
Since only a part of the semantics of the text is relevant to the frames $\mathbf{x}_{t-1}^k$ while obtained frames $\{M^i\}_{i=0}^{k-1}$ contain information about what has been generated, we propose to obtain frames-aware semantics $\mathbf{z}^{text}$ to guide motion generation. 

\hl{As shown in \Cref{fig:overview}.a}, the text is fed into a pretrained and fixed language model to obtain entire semantics. Then $L_1$ transformer decoder blocks \cite{vaswani2017attention} are used to acquire frames-aware semantics, where the \textit{query} is entire semantics, and \textit{key} and \textit{value} are both $\mathbf{z}^{k-1}$. \hl{Specifically, the cross-attention mechanism in our Frame-aware Semantics Decoder ensures that the original semantics, extracted from the pretrained language model, can quantitatively incorporate information from the given frames, thereby obtaining frame-aware semantics.}

\vspace{0.15cm}
\noindent\textbf{Text-Frame Guided Block.}
To generate $\mathbf{x}_{t-1}^k$ conditioned on frames-aware semantics and obtained frames, we propose a Text-Frame Guided Block (TGB) for generation.

\hl{As shown in \Cref{fig:overview}.b, in $l$-th block, embedded generating frames $\mathbf{z}^{k}_{l-1}$ are fed into a self-attention layer to explore the relation among generating frames, and into a cross-attention layer to explore the relation between generating frames and obtained frames.} Then, a Fusion Module is designed to fuse  outputs:
\begin{gather}
    \bar{\mathbf{z}}^{k}_l = LN(\mathbf{z}^{k}_{l-1} + Fusion(\mathbf{p}, \mathbf{q})), \label{eq:fusion} \\
    \mathbf{p}=SA(\mathbf{z}^{k}_{l-1}), \; \mathbf{q}=CA(\mathbf{z}^{k}_{l-1}, \mathbf{z}^{k-1}),
\end{gather}

\noindent
where $\mathbf{z}^{k}_0$ = $\mathbf{z}^{k}$, $LN(\cdot)$ is LayerNorm, and $SA(\cdot)$ and $CA(\cdot)$ denote the self-attention and cross-attention. Finally, $\bar{\mathbf{z}}^{k}_l$ and frames-aware semantics $\mathbf{z}^{text}$ are fed into another cross-attention layer to guide generation.

\hl{In detail, the Fusion Module is designed to fuse the outputs of the self-attention and the cross-attention, as shown in \Cref{eq:fusion}. As shown in \Cref{fig:overview}.c}, the concatenation of $\mathbf{p}$ and $\mathbf{q}$ is fed to an attention layer implemented by two fully-connected layers to obtain fusion weights $\mathbf{w}$. Then, the fusion is defined as:
\begin{equation}
    Fusion(\mathbf{p}, \mathbf{q}) = \mathbf{w} \odot \mathbf{p} + (\mathbf{1} - \mathbf{w}) \odot \mathbf{q},
\end{equation}

\noindent
where $\odot$ denotes the element-wise product.

\begin{algorithm}[t]
    \caption{Training Scheme with Pseudo-frames Replacement Strategy}
    \renewcommand{\algorithmicrequire}{\textbf{Requirement:}}
    \renewcommand{\algorithmicensure}{\textbf{Output:}}
    \begin{algorithmic}[1]
        {\small
        \REQUIRE Number of stages $K$; number of given frames $N_f$; number of diffusion steps $T$; frame replacement probability $\tau$; text replacement probability $\eta$; our PMG $g_\theta()$ and EMA PMG $g_\theta^{ema}()$;
        \WHILE{until required iterations }
             \STATE Sample a paired motion $M$ and text $\{w_i\}$ from the training set; 
             \STATE Sample $k$ from $\{1, ..., K\}$, $n_f$ from $\{1, ..., N_f\}$; 
             \STATE Set $n_f$ to $0$ with probability $\zeta$; 
             \STATE Sample $n_f$ frames from $M$ as given frames $\{f_i\}$;
            \STATE Divide the frames of $M$ into $K$ groups using \Cref{eq:divide1} and \Cref{eq:divide}; 
            \IF{$r_1 \sim U(0, 1) < \tau$ and $k > 1$} 
            \STATE Replace $\{M^i\}_{i=1}^{k-1}$ with  $\{\hat{M}^i\}_{i=1}^{k-1}$ predicted by $g_\theta^{ema}()$; 
            \ENDIF 
            \STATE Sample step $t$ from $\{1, ..., T\}$ and noise $\epsilon \sim \mathcal{N}(\textbf{0},\mathbf{I})$; 
            \STATE Obtain noisy frames $\mathbf{x}_t^k$ by adding noise to $M^k$: $\mathbf{x}_t^k = \text{AddNoise}(M^k, \epsilon, t)$; 
            \IF{$r_2 \sim U(0, 1) < \eta$} 
            \STATE Set $\{w_i\}$ to $\varnothing$; 
            \ENDIF 
            \STATE $\mathbf{c}^k = \{w_i\} \cup \{M^i\}_{i=1}^{k-1} \cup \{P^i\}_{i=0}^{k}$; 
            \STATE Predict noise $\epsilon_\theta(\mathbf{x}_t^k, t, \mathbf{c}^k)$ and compute loss via \Cref{eq:cfg2};
            \STATE Optimize $g_\theta()$ using the computed loss and update $g_\theta^{ema}()$; 
        \ENDWHILE
        }
    \label{algo}

    \end{algorithmic}
\end{algorithm}

\subsection{Training Strategy.}
\label{sec:training}
Although we assume the given frames are relevant to the text, providing meaningful postures, we still hope that our method will be more user-friendly and robust when given different frames during testing. Thus, we randomly select a few frames from the real motion as given frames and divide all frames into $K$ groups according to \Cref{eq:divide}. An integer $k$ is randomly sampled to determine the current stage. \Cref{algo} illustrates our detailed training scheme. Pseudo-code of the motion generation algorithm is provided in the supplementary material.

\vspace{0.15cm}
\noindent\textbf{Pseudo-frames Replacement Strategy.} 
Since frames generated in previous stages are not guaranteed to be correct, there is a discrepancy between training and testing, leading to errors accumulating over multiple stages. Hence, we propose a pseudo-frame replacement strategy to alleviate this discrepancy. We randomly replace frames $\{M^i\}_{i=1}^{k-1}$ with those generated by a momentum-based moving average of our generator (EMA model) during training.  In detail, we use the EMA model to predict $\{M^i\}_{i=1}^{k-1}$. The replacement probability $\tau$ is a hyper-parameter. In this way, our PMG is able to generate more accurate motion even when frames generated in previous stages are not entirely accurate.

\vspace{0.15cm}
\noindent\textbf{Generation Compatibility without Given Frames.}
To enhance the user-friendliness of our model, we introduce a simple method to enable generation without given keyframes. Specifically, during training, the number of given frames is set to 0 with a probability of $\zeta$, allowing the model to learn to generate motions without any given frames. Consequently, users can obtain an exemplar motion based solely on text conditions and can then design several desired postures.

\section{Experiments}

\subsection{Datasets} \label{sec:dataset}

\noindent\textbf{HumanML3D.} 
HumanML3D \cite{t2m} is the largest text-to-motion dataset that contains 14,616 3D human motion sequences and 44,970 text descriptions. The average motion length is 7.1 seconds, and the average length of text descriptions is 12 words. We use the same train-test split as \cite{t2m}.

\vspace{0.15cm}
\noindent\textbf{KIT Motion-Language (KIT-ML).}
KIT-ML \cite{kit} contains 3,911 3D human motion sequences and 6,278 text descriptions. The average motion length is 10.33 seconds, and the average length of text descriptions is 8.43 words. We use the same train-test split as \cite{tm2t, t2m}.

\vspace{0.15cm}
\noindent\textbf{Human Rating (HR).}
HR \cite{best-metric} collects 1,400 human motions generated by recent T2M models and annotates each text-motion pair with two human ratings (\textit{Naturalness} and \textit{Faithfulness}). HR is a small dataset used only to evaluate metrics but not for training.

\subsection{Implementaion Details} \label{sec:implent}
The number of stages $K$ is set to 3. The number of diffusion steps $T$ is 1000 during training, and the linearly varying variances $\beta_t$ range from $10^{-4}$ to 0.02. \la{We generate motion via UniPC \cite{zhao2023unipc}, and $T$ is set to 10 during generation to speed up the process.} The $d$, $L_1$, and $L_2$ are 512, 4, and 4, respectively. The pretrained language model is CLIP \cite{clip}. \la{We trained our PMG with a batch size of 512 on two RTX 3090 GPUs. For the HumanML3D dataset, we trained for 3.2K epochs, and for the KIT-ML dataset, we trained for 16K epochs.} The Adam optimizer is used with a learning rate of $10^{-4}$. $\tau$, $s$, $\eta$, and $\zeta$ are set to 0.3, 2, 0.2, and 0.001, respectively. \la{Additionally, during testing, the absolute position information of the root joint is discarded to align with the motion representation used by the previous T2M evaluator \cite{t2m}.}

\subsection{Evaluation Metrics}
Following \cite{t2m}, we adopt the T2M-Evaluator \cite{t2m} to obtain the feature embeddings of motion and text, and then use these to compute the following metrics: 1) Frechet Inception Distance (FID) evaluates the distance between the generated motions and real motions. 2) R-Precision calculates the motion-retrieval precision when given one motion and 32 texts (1 real text and 31 randomly sampled mismatched texts). 3) Multimodal Distance (MM-Dist) evaluates the distance between motions and texts. 4) Diversity evaluates the diversity of generated motions.

Moreover, \cite{best-metric} shows that coordinate error metrics are also important. As suggested by \cite{best-metric}, we adopt Average Variance Error (AVE) \cite{hier}, Average Position Error (APE) \cite{Language2Pose}, and MoBERT Score \cite{best-metric} for evaluation.

\begin{figure}[t]
    \centering
    \includegraphics[width=\linewidth]{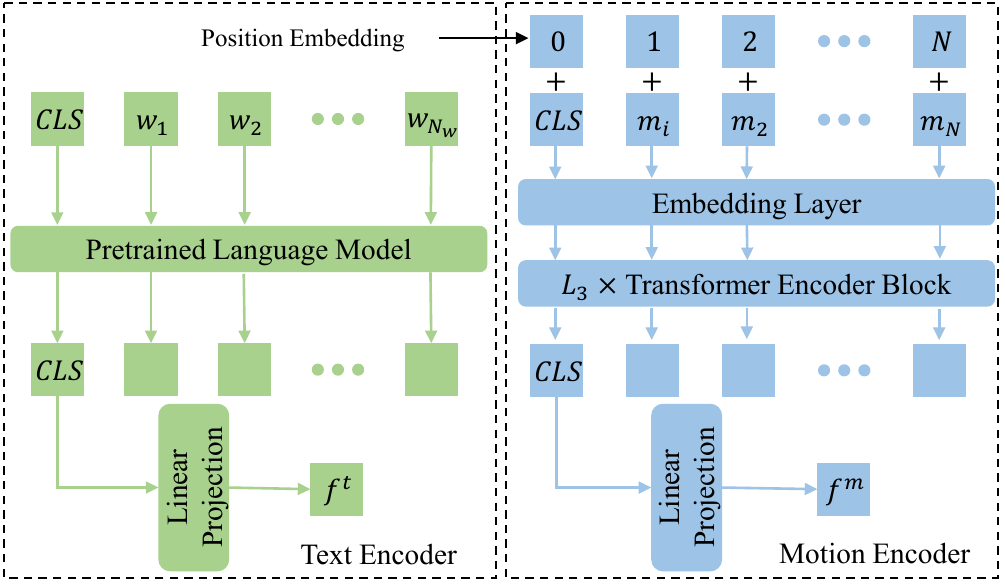}
    \caption{\hl{The illustration of our MotionCLIP, which maps matched text-motion pairs into closely aligned feature vectors in the latent space. MotionCLIP consists of two main components: a motion encoder and a text encoder.}}
    \label{fig:motionclip}
\end{figure}

\begin{table}[]
    \centering
    \setlength\tabcolsep{3pt}
    \caption{Pearson correlations with human judgments on HR \cite{best-metric} dataset. A higher absolute value indicates a higher correlation with human judgments. The best result is in bolded.}
    \begin{tabular}{ccccc}
    \toprule[1pt]
           &   \multicolumn{2}{c}{R-Precision}  & \multicolumn{2}{c}{FID} \\ \cmidrule(lr){2-3} \cmidrule(lr){4-5}
           & Faithfulness & Naturalness & Faithfulness & Naturalness \\ 
    \midrule
        \makecell[c]{T2M \\ Evaluator\cite{t2m}} & 0.816\cite{best-metric} & 0.756\cite{best-metric} & -0.714\cite{best-metric} & -0.269\cite{best-metric} \\ 
        \makecell[c]{Our \\ MotionCLIP} & \textbf{0.961} & \textbf{0.762} & \textbf{-0.989} & \textbf{-0.801} \\ 
    \bottomrule[1pt]
    \end{tabular}
    \label{tab:hr}
\end{table}

\vspace{0.1cm}
\noindent\textbf{MotionCLIP.} 
We provide a strong evaluator to assess the quality of generated motions, since it is hard for the T2M-Evaluator \cite{t2m} to distinguish between real motions and motions generated by recent methods. As shown in \Cref{fig:motionclip}, our MotionCLIP, consisting of a motion encoder and a text encoder, is trained with the contrastive loss in CLIP \cite{clip}. More details are provided in the supplementary material. 

\begin{table*}[t]
    \centering
    \caption{Quantitative results on HumanML3D\cite{t2m} test set. Following T2M\cite{t2m}, we repeat the evaluation \textbf{20} times and report the average with a $95\%$ confidence interval. We evaluate the released checkpoints of some works using new metrics, denoted by $*$. $\bullet$ denotes that this model is trained on the TF2M setting. The best result is in bold, the second best result is underlined. $\rightarrow$ indicates that values closer to those of real motion are better.}
    \setlength\tabcolsep{3pt}
    \begin{tabular}{cccccccccc}
        \toprule[1pt]
        \multirow{2}{*}{\makecell{\#Given\\frames}} & \multicolumn{1}{c}{\multirow{2}{*}{Methods}} & \multicolumn{2}{c}{MotionCLIP} & \multicolumn{4}{c}{T2M Evaluator\cite{t2m}} & \multicolumn{1}{c}{\multirow{2}{*}{\makecell{MoBert \cite{best-metric} \\ Alignment$\uparrow$}}} & \multicolumn{1}{c}{\multirow{2}{*}{AVE$^{Root}\downarrow$}} \\ \cmidrule(lr){3-4} \cmidrule(lr){5-8}
         & \multicolumn{1}{c}{} & R-Top1$\uparrow$ & \multicolumn{1}{c}{FID$^\star\downarrow$}  
                & R-Top1$\uparrow$ & FID$\downarrow$ & MM-Dist$\downarrow$ & \multicolumn{1}{c}{Diversity$\rightarrow$} 
                & \multicolumn{1}{c}{}   \\ \midrule
        - & Real motion & $0.752^{\pm.002}$ & $0.001^{\pm.000}$ & $0.511^{\pm.003}$ & $0.002^{\pm.000}$ & $2.974^{\pm.008}$ & $9.503^{\pm.065}$ & $0.621^{\pm.002}$ & 0 \\ \hline 
        \multirow{9}{*}{0} & T2M\cite{t2m} & - & - & $0.455^{\pm.003}$ & $1.087^{\pm.021}$ & $3.347^{\pm.008}$ & $9.175^{\pm.083}$ & - & -\\
         & MDM\cite{mdm} & - & - & $0.320^{\pm.005}$ & $0.544^{\pm.044}$ & $5.566^{\pm.027}$ & $\underline{9.559}^{\pm.086}$
                & - & - \\
         & MLD\cite{mld} & - & - & $0.481^{\pm.003}$ & $0.473^{\pm.013}$ & $3.196^{\pm.010}$ & $9.724^{\pm.082}$
                & - & -\\
        & Fg-T2M\cite{fgt2m} & - & - & $0.492^{\pm.002}$ & $0.243^{\pm.019}$ & $3.109^{\pm.007}$ & $9.278^{\pm.072}$
                & - & -\\
         & MotionDiffuse\cite{md} & $0.665^{\pm.002}_*$ & $3.996^{\pm.040}_*$ & $0.491^{\pm.001}$ & $0.630^{\pm.001}$ & $3.113^{\pm.001}$ & $9.410^{\pm.049}$ & $0.561^{\pm.003}_*$ & $0.197^{\pm.043}_*$ \\
         & T2M-GPT\cite{t2m-gpt} & $0.661^{\pm.003}_*$ & $1.386^{\pm.023}_*$ & $0.491^{\pm.003}$ & $0.116^{\pm.004}$ & $3.118^{\pm.011}$ & $9.761^{\pm.081}$ & $0.597^{\pm.003}_*$ & $0.179^{\pm.032}_*$\\
         & ReMoDiffuse\cite{zhang2023remodiffuse} & $0.661^{\pm.003}_*$ & $1.981^{\pm.020}_*$ & $0.510^{\pm.005}$ & $0.103^{\pm.004}$ & $2.974^{\pm.016}$ & $9.018^{\pm.075}$ & $0.589^{\pm.002}_*$ & $0.157^{\pm.026}_*$ \\
         & MoMask\cite{momask} & $0.722^{\pm.003}_*$ & $1.344_*^{\pm.021}$ & $0.521^{\pm.002}$ & $0.045^{\pm.002}$ & $2.958^{\pm.008}$ & $9.685_*^{\pm.087}$ & $0.621_*^{\pm.002}$ & $0.122_*^{\pm.064}$ \\ 
         & BAMM\cite{bamm} & $0.729^{\pm.004}_*$ & $1.369_*^{\pm.019}$ & $0.522^{\pm.003}$ & $0.055^{\pm.002}$ & $2.936^{\pm.077}$ & $9.636^{\pm.087}$ & $0.625_*^{\pm.005}$ & $0.124_*^{\pm.047}$ \\ 
         \hline
          & ReMoDiffuse$^\bullet$\cite{zhang2023remodiffuse} & $0.663^{\pm.002}$ & $1.923^{\pm.030}$ & $0.508^{\pm.003}$ & $0.105^{\pm.002}$ & $3.035^{\pm.012}$ & $9.248^{\pm.010}$ & $0.593^{\pm.003}$ & $0.158^{\pm.031}$\\
          & OmniControl$^\bullet$\cite{omnicontrol} & $0.598^{\pm.004}$ & $2.681^{\pm.051}$ & $0.434^{\pm.005}$ & $0.461^{\pm.031}$ & $3.186^{\pm.018}$ & $9.376^{\pm.046}$ & $0.565^{\pm.007}$ & $0.195^{\pm.037}$\\
           & BAMM$^\bullet$\cite{bamm} & $0.721^{\pm.004}$ & $1.376^{\pm.024}$ & $0.519^{\pm.004}$ & $0.057^{\pm.003}$ & $2.947^{\pm.061}$ & $9.621^{\pm.076}$ & $0.624^{\pm.003}$ & $0.120^{\pm.035}$ \\ 
          \rowcolor{orange!30} \cellcolor{white} \multirow{-4}{*}{1} & Our PMG & $\underline{0.757}^{\pm.002}$ & $\underline{0.401}^{\pm.004}$ & $\underline{0.535}^{\pm.003}$ & $\underline{0.022}^{\pm.001}$ & $\underline{2.834}^{\pm.005}$ & $9.560^{\pm.092}$ 
            & $\underline{0.646}^{\pm.002}$  & $\underline{0.080}^{\pm.001}$\\ \hline
          & ReMoDiffuse$^\bullet$\cite{zhang2023remodiffuse} & $0.689^{\pm.002}$ & $1.582^{\pm.020}$ & ${0.511}^{\pm.005}$ & $0.095^{\pm.005}$ & $2.964^{\pm.008}$ & $9.197^{\pm.054}$ & $0.598^{\pm.003}$ & $0.150^{\pm.033}$ \\
          & OmniControl$^\bullet$\cite{omnicontrol} & $0.622^{\pm.002}$ & $2.135^{\pm.043}$ & $0.467^{\pm.003}$ & $0.265^{\pm.012}$ & $3.115^{\pm.019}$ & $9.247^{\pm.076}$ & $0.573^{\pm.005}$ & $0.186^{\pm.039}$\\
          & BAMM$^\bullet$\cite{bamm} & $0.737^{\pm.001}$ & $1.210^{\pm.018}$ & $0.526^{\pm.002}$ & $0.043^{\pm.002}$ & $2.924^{\pm.039}$ & $9.592^{\pm.063}$ & $0.631^{\pm.002}$ & $0.102^{\pm.034}$ \\ 
          \rowcolor{orange!30} \cellcolor{white} \multirow{-4}{*}{2}   &  Our PMG & $\pmb{0.763}^{\pm.003}$ & $\pmb{0.350}^{\pm.004}$ & $\pmb{0.536}^{\pm.002}$ & $\pmb{0.018}^{\pm.001}$ & $\pmb{2.832}^{\pm.009}$ & $\pmb{9.528}^{\pm.086}$ 
            & $\pmb{0.650}^{\pm.002}$ & $\pmb{0.068}^{\pm.001}$ \\
         \bottomrule[1pt]
    \end{tabular}
    \label{tab:hml3d}
\end{table*}

\la{
\subsection{Quantitative Evaluation of Our MotionCLIP} \label{sec:results}
As shown in the first row of \Cref{tab:hml3d}, our MotionCLIP outperforms the T2M-Evaluator (T2M-E) \cite{t2m} by a large margin (0.752 vs. 0.511). \hl{Recent works such as T2M-GPT \cite{t2m-gpt}, MoMask \cite{momask}, and BAMM \cite{bamm}} achieve similar results to real motions when evaluated by T2M-E, showing that T2M-E is not able to evaluate recent works accurately. Additionally, our MotionCLIP is also evaluated on the HR \cite{best-metric} dataset and achieves significant improvements, as shown in \Cref{tab:hr}. These results demonstrate that our MotionCLIP is better than T2M-E at evaluating generated motions.

\subsection{Quantitative Evaluation of Our PMG}
\noindent\textbf{Comparative Methods.}
Since our work is the first to explore the text-frame-to-motion (TF2M) generation task, in addition to comparing with previous text-to-motion (T2M) generation methods, \hl{we also construct TF2M baselines based on below considerations for a fair comparison:}
\hl{
\begin{itemize}
    \item For diffusion-based T2M methods, \ie, ReMoDiffuse \cite{zhang2023remodiffuse}, that generate motion in the original motion representation space, it is easy to replace some frames with given frames during training or testing to incorporate the TF2M generation task.
    \item For VAE-based T2M methods, such as BAMM \cite{bamm}, which generate motion in the latent space rather than in the original motion representation space, we modify the motion generation modules to incorporate conditioning on provided frames. Specifically, since BAMM is a transformer-based model that autoregressively predicts motion tokens conditioned on a text embedding prompt, the modified BAMM$^\bullet$ leverages both the text embedding and the given frames—rather than relying solely on the text embedding—to guide the generation of subsequent motion tokens, thereby fulfilling the requirements of the TF2M generation task.
    \item For controllable motion generation methods such as OmniControl \cite{omnicontrol} that incorporate key joint positions, the key joint positions can be replaced with the full-body human joint information from the given frames to incorporate the TF2M generation task.
\end{itemize}
}

\hl{Based on the above considerations, we construct three TF2M baselines, \ie, ReMoDiffuse$^\bullet$, OmniControl$^\bullet$, BAMM$^\bullet$, using the modification method mentioned above. }
The training and testing settings for key frames are consistent with those of our model. To evaluate the approximate average performance when given frames are meaningful or meaningless, we randomly select given frames from the ground-truth motion.

\vspace{0.2cm}
\noindent\textbf{Results and Analysis.}
\Cref{tab:hml3d} and \Cref{tab:kit} show the important metrics of all methods, and full results are shown in the supplementary material. \hl{Note that the method using 0 given frames is the T2M method, while all other methods (denoted by the symbol $^\bullet$) are referred to as TF2M methods.}

As shown in both tables, even with only one given frame, our PMG still significantly outperforms existing works, especially in FID and AVE$^{Root}$. Our PMG achieves 0.022/0.118 and 0.401/1.258 on T2M-FID and MotionCLIP-FID$^\star$\footnote{ \footnotesize{FID$^\star$ indicates the values are multiplied by 100 for best display.}} on both datasets when given one frame, outperforming existing methods by a large margin. We find that more given frames lead to lower FID and AVE$^{Root}$. \hl{ On the other hand, both results of ReMoDiffuse$^\bullet$ and BAMM$^\bullet$ indicate that simply inputting clean frames into a diffusion model—or using these frames as prompts for an autoregressive model—does not yield significant improvements. } Moreover, the results of OmniControl$^\bullet$ indicate that existing controllable T2M methods designed for controlling key joint positions are not capable of performing the text-frame-to-motion generation task.

Furthermore, we find that our PMG and some methods achieve higher results than real motions on some metrics, probably because T2M-E \cite{t2m} and MotionCLIP are not strong enough to distinguish between generated motions and real motions. Another possible reason is that the generated motions are overfitting on the text but lose realism\footnote{One piece of evidence is that MotionDiffuse \cite{md} and Fg-T2M \cite{fgt2m} achieve higher R-Top1 but worse FID, MoBert Alignment, and AVE$^{Root}$.}. However, our PMG achieves not only higher R-Top1 but also better FID and AVE$^{Root}$, demonstrating that our PMG generates more realistic motions.

\vspace{0.2cm}
\noindent\textbf{Discussion.} As shown in \Cref{tab:hml3d} and \Cref{tab:kit}, although MotionCLIP is stronger than T2M-E \cite{t2m}, it is still not strong enough to distinguish between generated motions and real motions. The newly proposed MoBert \cite{best-metric} faces the same issue. These findings remind us that we need to combine multiple metrics to evaluate generated motions, and that we urgently need better methods to evaluate generation models.
}

\begin{table*}[]
    \centering
    \caption{Quantitative results on KIT-ML\cite{kit} test set. Following T2M\cite{t2m}, we repeat the evaluation \textbf{20} times and report the average with a $95\%$ confidence interval. We evaluate the released checkpoints of some works using new metrics, denoted by $*$. $\bullet$ denotes that this model is trained on the TF2M setting. The best result is in bold, the second best result is underlined. $\rightarrow$ indicates that values closer to those of real motion are better.}
    \setlength\tabcolsep{3pt}
    \begin{tabular}{ccccccccc}
        \toprule[1pt]
        \multirow{2}{*}{\makecell{\#Given\\frames}} & \multicolumn{1}{c}{\multirow{2}{*}{Methods}} &   \multicolumn{2}{c}{MotionCLIP} & \multicolumn{4}{c}{T2M Evaluator\cite{t2m}} & \multicolumn{1}{c}{\multirow{2}{*}{AVE$^{Root}\downarrow$}}\\ \cmidrule(lr){3-4} \cmidrule(lr){5-8}
         & \multicolumn{1}{c}{} & R-Top1$\uparrow$ & \multicolumn{1}{c}{FID$^\star\downarrow$}  
                &  R-Top1$\uparrow$ & FID$\downarrow$ & MM-Dist$\downarrow$ & \multicolumn{1}{c}{Diversity$\rightarrow$} 
                & \multicolumn{1}{c}{}   \\ \midrule
        - &  \textbf{Real motion} & $0.536^{\pm.006}$ & $0.013^{\pm.005}$ & $0.424^{\pm.005}$ & $0.031^{\pm.004}$ & $2.788^{\pm.012}$ & $11.08^{\pm.097}$ & 0 \\ \hline
        \multirow{8}{*}{0} & T2M\cite{t2m} & - & - & $0.361^{\pm.006}$ & $3.022^{\pm.107}$ & $3.488^{\pm.028}$ & $10.72^{\pm.145}$ & - \\
         & MDM\cite{mdm} & - & - & $0.164^{\pm.005}$ & $0.497^{\pm.021}$ & $9.191^{\pm.022}$ & $10.85^{\pm.109}$ & - \\
         & MLD\cite{mld} & - & - & $0.390^{\pm.008}$ & $0.404^{\pm.027}$ & $3.204^{\pm.027}$ & $10.80^{\pm.117}$ & - \\
          & Fg-T2M\cite{fgt2m} & - & - & $0.418^{\pm.005}$ & $0.571^{\pm.047}$ & $3.114^{\pm.015}$ & $10.93^{\pm.083}$ & - \\
         & MotionDiffuse\cite{md} & $0.567^{\pm.004}_*$ & $7.281^{\pm.085}_*$ & $0.417^{\pm.004}$ & $1.954^{\pm.064}$ & $2.958^{\pm.005}$ & $\underline{11.10}^{\pm.143}$ &  $0.496^{\pm.101}_*$ \\
         & T2M-GPT\cite{t2m-gpt} & $0.537^{\pm.005}_*$ & $2.703^{\pm.055}_*$ & $0.416^{\pm.006}$ & $0.514^{\pm.029}$ & $3.007^{\pm.023}$ & $10.92^{\pm.108}$ & $0.443^{\pm.060}_*$\\
         & ReMoDiffuse\cite{zhang2023remodiffuse} & $0.550^{\pm.006}_*$ & $2.332^{\pm.031}_*$ & $0.427^{\pm.014}$ & ${0.155}^{\pm.006}$ & $2.814^{\pm.012}$ & $10.80^{\pm.105}$ & $0.453^{\pm.100}_*$\\
         & MoMask\cite{momask} & $0.568^{\pm.006}_*$ & $2.699^{\pm.016}_*$ & $0.433^{\pm.007}$ & ${0.204}^{\pm.011}$ & $2.779^{\pm.022}$ & $10.82^{\pm.091}_*$ & $0.439^{\pm.057}_*$\\
            \hline
        &  ReMoDiffuse$^\bullet$\cite{zhang2023remodiffuse} & $0.531^{\pm.007}$ & $2.491^{\pm.040}$ & $0.426^{\pm.005}$ & $0.169^{\pm.010}$ & $2.860^{\pm.020}$ & $10.75^{\pm.086}$ & $0.451^{\pm.069}$\\
        & OmniControl$^\bullet$\cite{omnicontrol} & $0.541^{\pm.005}$ & $2.795^{\pm.072}$  & $0.387^{\pm.007}$ & $0.632^{\pm.021}$ &  $3.115^{\pm.019}$ & $10.84^{\pm.051}$ & $0.468^{\pm.058}$\\
        \rowcolor{orange!30} \cellcolor{white} \multirow{-3}{*}{1}  & Our PMG & 
            $\pmb{0.612}^{\pm.007}$ & $\underline{1.258}^{\pm.012}$ & $\pmb{0.453}^{\pm.005}$ & $\underline{0.118}^{\pm.004}$ & $\underline{2.637}^{\pm.015}$ & ${11.11}^{\pm.217}$ & $\underline{0.323}^{\pm.003}$ \\ \hline
        & ReMoDiffuse$^\bullet$\cite{zhang2023remodiffuse} & $0.562^{\pm.006}$ & $2.260^{\pm.026}$ & $0.429^{\pm.006}$ & $0.152^{\pm.001}$ & $2.771^{\pm.016}$ & $10.74^{\pm.080}$ & $0.431^{\pm.081}$\\
        & OmniControl$^\bullet$\cite{omnicontrol} &  $0.549^{\pm.004}$ & $2.651^{\pm.041}$ & $0.400^{\pm.006}$ & $0.565^{\pm.026}$ & $3.085^{\pm.025}$ & $10.67^{\pm.072}$ &   $0.462^{\pm.056}$\\
        \rowcolor{orange!30} \cellcolor{white} \multirow{-3}{*}{2}  & Our PMG & $\underline{0.603}^{\pm.007}$ & $\pmb{1.119}^{\pm.012}$ & $\underline{0.451}^{\pm.007}$ & $\pmb{0.101}^{\pm.002}$ & $\pmb{2.630}^{\pm.016}$ & $\pmb{11.09}^{\pm.127}$ & $\pmb{0.308}^{\pm.003}$ \\
         \bottomrule[1pt]
    \end{tabular}
    \label{tab:kit}
\end{table*}

\begin{figure*}[]
    \centering
    \includegraphics[width=\linewidth]{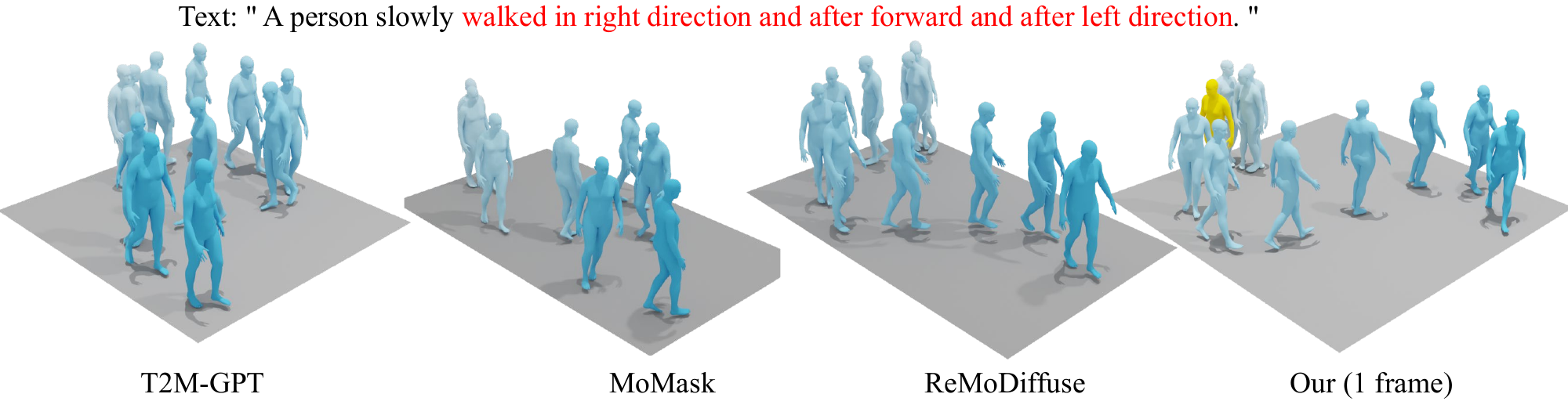}
    \caption{Visual comparisons on HumanML3D \cite{t2m} dataset. The test sample is \#013150. We visualize the generated motion of PMG when given one frame (indicated in yellow). The motion generated by PMG is more consistent with the given text.}
    \label{fig:visual1}
\end{figure*}

\begin{figure*}[]
    \centering
    \includegraphics[width=\linewidth]{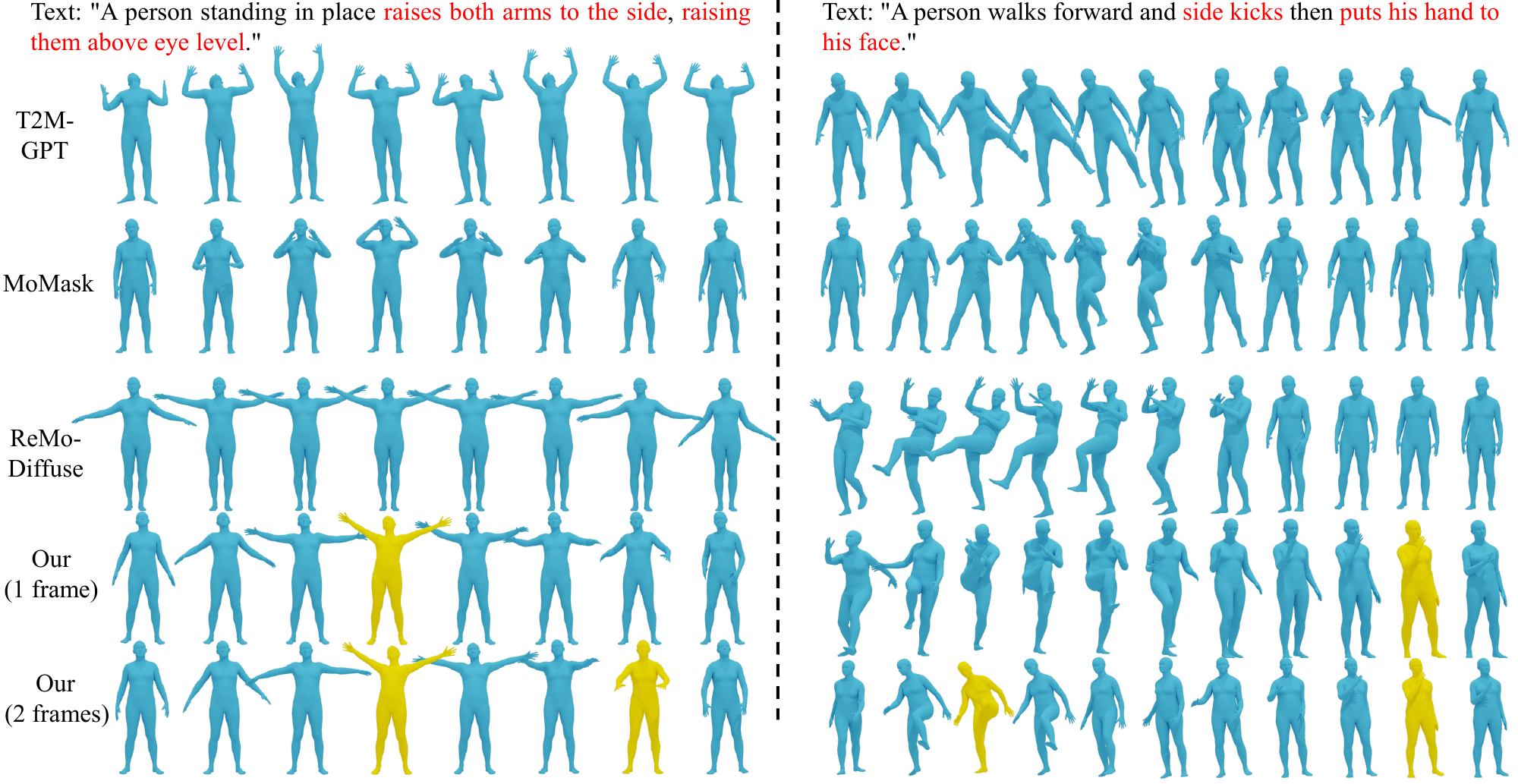}
    \caption{Visual comparisons on HumanML3D \cite{t2m} dataset. The test samples are \#000534 and \#010797. Note that these two samples are hard samples. We visualize the generated motions of our PMG when given different frames (indicated in yellow). }
    \label{fig:visual2}
\end{figure*}

\subsection{Qualitative Evaluation} \label{sec:visual}

\noindent\textbf{Visual Comparisons with Previous Methods.} 
In \Cref{fig:visual1}, only the motions generated by our PMG and ReMoDiffuse\cite{zhang2023remodiffuse} are consistent with the given text, and the motion generated by our PMG is even more aligned with the text. Moreover, \Cref{fig:visual2} shows examples of two challenging cases. Previous methods struggle to generate realistic and smooth motions consistent with the given text. For example, motions generated by ReMoDiffuse\cite{zhang2023remodiffuse} are incomplete and look somewhat strange.
By contrast, even with only one given frame, our method can generate realistic and smooth motions. With more given frames, the generated motion becomes even more realistic.

\begin{figure*}[t]
    \centering
    \includegraphics[width=\linewidth]{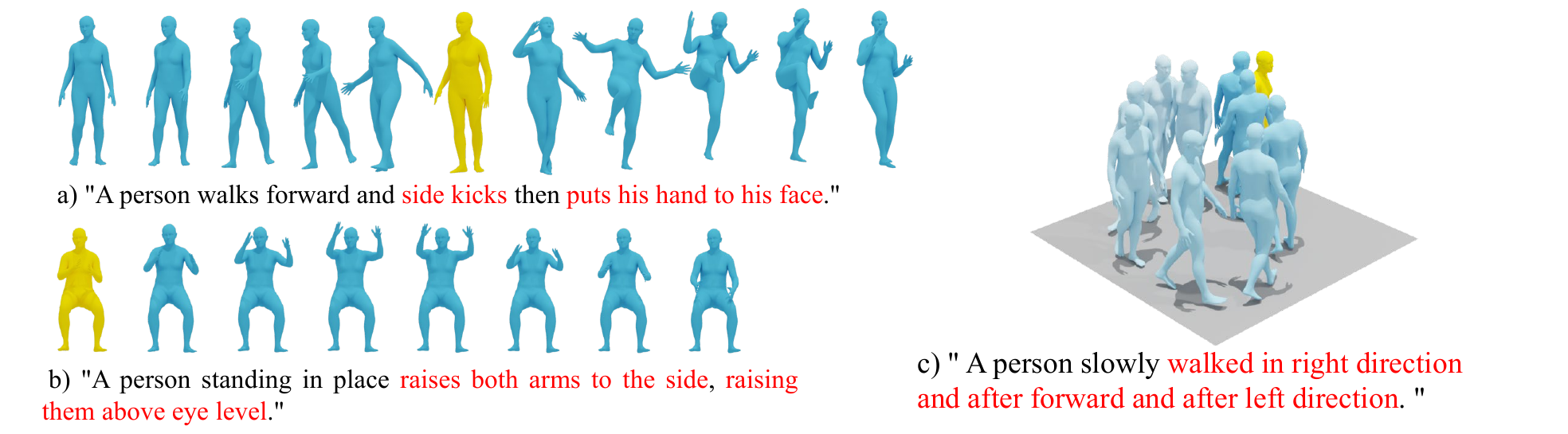}
    \caption{Visual results of our PMG when given low-quality frames. (a) The given frame is meaningless and in the middle. (b) The given frame is not relevant to the text (a sitting posture). (c) The given frame conflicts with the text (the human's position does not match the text). }
    \label{fig:visual3}
\end{figure*}

\begin{table*}[t]
    \centering
    \caption{Experimental results on HumanML3D-Sub with different given frames. ``Rand./Anno.'' denotes using randomly selected human frames or annotated frames as given frames for testing.   Note that FID is larger on HumanML3D-Sub than on HumanML3D due to the number of test samples. } 
    \begin{tabular}{ccccccc}
        \toprule[1pt]
        \multirow{2}{*}{\makecell{\#Given\\frames}} &  \multirow{2}{*}{Methods} & \multicolumn{2}{c}{MotionCLIP} &   \multicolumn{2}{c}{T2M Evaluator\cite{t2m}} & \multicolumn{1}{c}{\multirow{2}{*}{AVE$^{Root}\downarrow$}} \\ \cmidrule(lr){3-4} \cmidrule(lr){5-6}
        & &  Top-1$\uparrow$ & \multicolumn{1}{c}{FID$^\star\downarrow$} & Top-1$\uparrow$ & \multicolumn{1}{c}{FID$\downarrow$}   \\ \midrule
        \multirow{2}{*}{0} & ReMoDiffuse \cite{zhang2023remodiffuse} & ${0.664}^{\pm.005}_*$ & ${3.492}^{\pm.050}_*$ & ${0.507}^{\pm.005}_*$ & ${0.196}^{\pm.016}_*$ & ${0.202}^{\pm.048}_*$\\ 
         & BAMM \cite{bamm} & ${0.721}^{\pm.005}_*$ & ${2.156}^{\pm.058}_*$ & ${0.524}^{\pm.006}_*$ & ${0.124}^{\pm.007}_*$ & ${0.142}^{\pm.044}_*$\\ 
        \hline
        & ReMoDiffuse$^\bullet$ \cite{zhang2023remodiffuse} & ${0.685}^{\pm.004}$ & ${3.385}^{\pm.080}$ & ${0.512}^{\pm.005}$ & ${0.183}^{\pm.009}$ & ${0.134}^{\pm.002}$ \\
        & BAMM$^\bullet$ \cite{bamm} & ${0.732}^{\pm.004}$ & ${1.976}^{\pm.046}$ & ${0.529}^{\pm.004}$ & ${0.102}^{\pm.006}$ & ${0.118}^{\pm.029}$\\ 
        \rowcolor{orange!30} \cellcolor{white} \multirow{-3}{*}{\makecell{\cellcolor{white}2\\\cellcolor{white}{(Rand.)}}} &  \textbf{Our PMG} & $\underline{0.751}^{\pm.005}$ & $\underline{1.109}^{\pm.020}$ & $\underline{0.542}^{\pm.008}$ & $\underline{0.068}^{\pm.005}$ & $\underline{0.076}^{\pm.001}$  \\ \hline
         & ReMoDiffuse$^\bullet$ \cite{zhang2023remodiffuse} & ${0.694}^{\pm.006}$ & ${3.093}^{\pm.060}$ & ${0.515}^{\pm.007}$ & ${0.170}^{\pm.014}$ & $0.108^{\pm.019}$ \\
         & BAMM$^\bullet$ \cite{bamm} & ${0.738}^{\pm.003}$ & ${1.785}^{\pm.038}$ & ${0.534}^{\pm.002}$ & ${0.094}^{\pm.007}$ & ${0.097}^{\pm.017}$\\ 
        \rowcolor{orange!30} \cellcolor{white} \multirow{-3}{*}{\makecell{\cellcolor{white}2\\\cellcolor{white}{(Anno.)}}} &   \textbf{Our PMG} & $\pmb{0.755}^{\pm.004}$ & $\pmb{0.946}^{\pm.013}$ & $\pmb{0.547}^{\pm.007}$ & $\pmb{0.057}^{\pm.003}$ & $\pmb{0.064}^{\pm.001}$ \\ \bottomrule[1pt]
    \end{tabular}
    \label{tab:hml3d-sub}
\end{table*}

\vspace{0.3cm}
\noindent\textbf{Robustness When Given Low-quality Frames.}
\Cref{fig:visual3} shows that our PMG continues to generate text-relevant actions even when given low-quality frames, since our training strategy makes our PMG robust.  According to Figures. \Cref{fig:visual1}, \Cref{fig:visual2}, and \Cref{fig:visual3}, we can conclude that meaningful frames help our PMG generate realistic motions with desired postures, and that our PMG is robust to variations in the input frames.

\begin{figure}[t]
    \centering
    \includegraphics[width=0.8\linewidth]{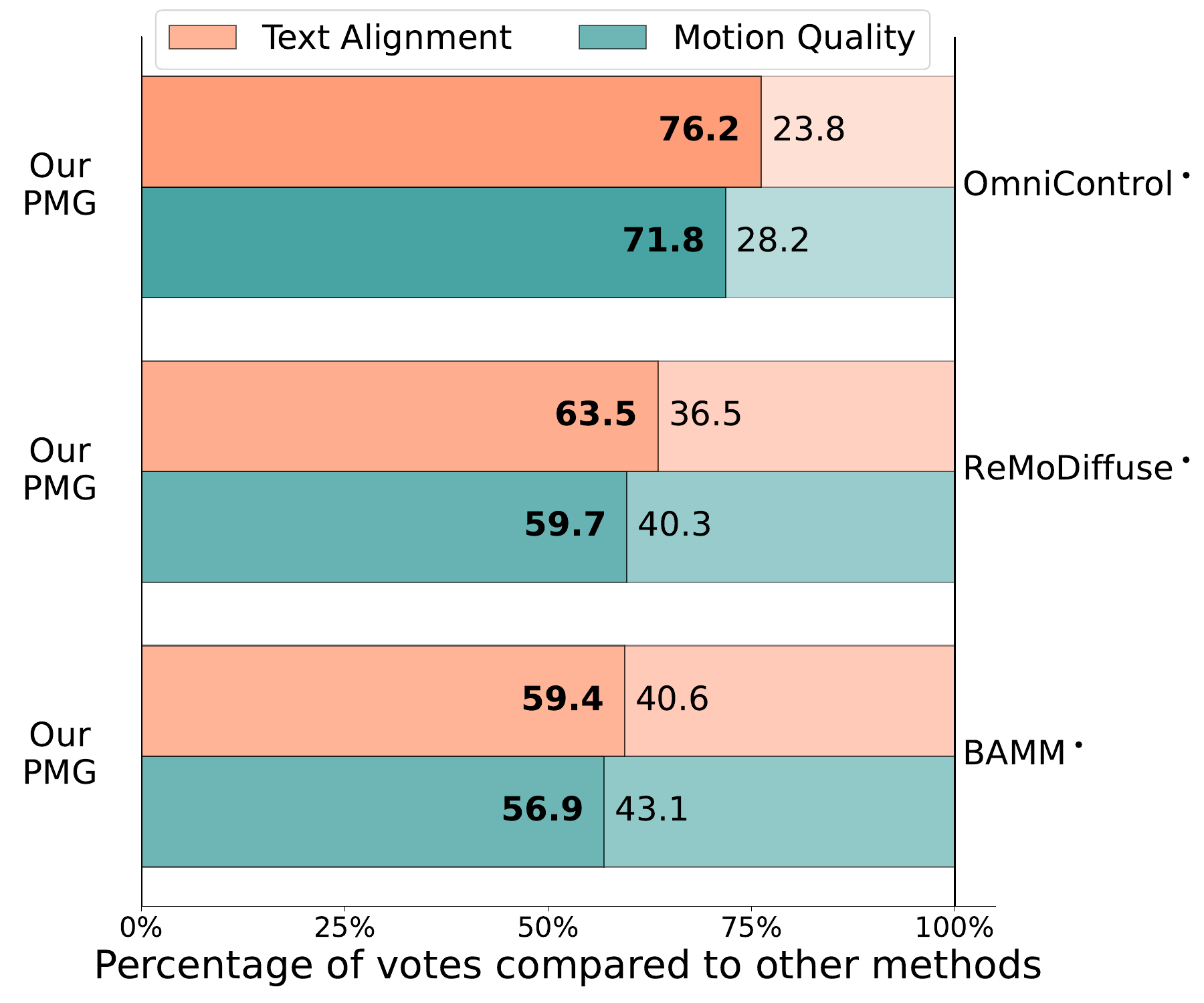}
    \caption{User Study. The color bar and numbers indicate the preference rate of our PMG over the compared methods.}
    \label{fig:user-study}
    \vspace{-0.3cm}
\end{figure}

\hl{
\vspace{0.1cm}
\noindent\textbf{User Study.} Furthermore, we conduct a user study to evaluate the effectiveness of our method. Specifically, we select 30 hard-to-generate test samples using the R-Top1 metric to assess the difficulty of generating a given text. Each method then generates three motions for each test sample. As a result, we obtain 90 generated motions for each method, which are then subjected to 1-on-1 comparisons. In side-by-side comparisons, 25 participants are asked to select the one that better aligns with the given text or shows higher motion quality. As shown in \Cref{fig:user-study}, we observe that although all samples are inherently challenging to generate (hard-to-generate samples), our method is preferred by users most of the time. Additionally, we note that our model demonstrates relatively stronger performance in text alignment compared to motion quality. Please refer to the Appendix for more details.

}

\subsection{Ablation Study} \label{sec:ablation}

All experiments below are conducted with two given frames on HumanML3D\cite{t2m} unless otherwise stated. Please refer to the Appendix for more experiments.

\vspace{0.15cm}
\noindent\textbf{Robustness of Our PMG.}
To validate the robustness of our PMG when given frames of different qualities, we collect a subset of the HumanML3D \cite{t2m} test set containing a quarter of its samples (1000 samples), named HumanML3D-Sub, and design two types of given frames for each test sample: 
\begin{itemize}
    \item \textit{Human-Annotated Frames.} For each text-motion sample, volunteers annotated a few frames from the ground-truth motion that best reflect the action posture described in the text as the given frames. These human-annotated frames, containing the most meaningful postures, are considered high-quality given frames. 
    \item \textit{Randomly Selected Human Frames.} For each text-motion sample, we randomly select a few frames from the ground-truth motion as the given frames. These randomly selected human frames, which may represent meaningful or meaningless postures, approximate different user input frames. 
\end{itemize}

\hl{We also evaluate the performance of both the unmodified methods (ReMoDiffuse \cite{zhang2023remodiffuse} and BAMM \cite{bamm}) and the modified methods (ReMoDiffuse$^\bullet$ and BAMM$^\bullet$ \cite{bamm}) on HumanML3D-Sub.} Note that all methods are trained on HumanML3D, and HumanML3D-Sub is used only for testing.

The results are shown in \Cref{tab:hml3d-sub}. Importantly, the results obtained when using human-annotated frames can serve as an upper bound for our method, while the results with randomly selected human frames can serve as a benchmark for average performance when the given frames are text-relevant but may not be particularly meaningful. Hence, the performance of our PMG with randomly selected human frames demonstrates its robustness and user-friendliness. The difference in performance between "Anno." and "Rand." indicates that our method achieves higher performance when the given frames are more meaningful. We also provide visualizations of the results with different frames in  \Cref{sec:visual}.

\begin{table}[]
    \centering
    \caption{Ablation studies on the generation stage $K$. We evaluate the inference time for test sample ``000021'' which contains 179 frames and report the average inference time over 100 tests. All tests are conducted on the same RTX 3090. }
    \setlength\tabcolsep{3pt}
    \begin{tabular}{ccccccc}
        \toprule[1pt]
        \multirow{2}{*}{} & \multicolumn{1}{c}{\multirow{2}{*}{\#Params}} & \multicolumn{1}{c}{\multirow{2}{*}{\makecell{Inference\\Time }}} &   \multicolumn{2}{c}{MotionCLIP} & \multicolumn{1}{c}{\multirow{2}{*}{AVE$^{Root}\downarrow$}}  \\ \cline{4-5}
         &  \multicolumn{1}{c}{} & \multicolumn{1}{c}{} & \multicolumn{1}{c}{R-Top1$\uparrow$} & \multicolumn{1}{c}{FID$^\star\downarrow$}  \\ \midrule
         MotionDiffuse\cite{md} & 86.00M & 16.82s & 0.665$_*$ & 3.996$_*$ & 0.197$_*$  \\
         T2M-GPT\cite{t2m-gpt} & 237.6M & 0.73s & 0.661$_*$ & 1.386$_*$ & 0.179$_*$ \\ 
         ReMoDiffuse\cite{zhang2023remodiffuse}
          & 44.14M & 0.91s & $0.661_*$ & $1.981_*$ & $0.157_*$ \\
         \hline
         K = 1 & 18.30M & \textbf{0.18s} & 0.743 & 0.510 & 0.078 \\ 
         K = 2 & 18.30M & 0.36s & \textbf{0.763} & 0.360 & 0.070 \\
         \rowcolor{orange!30} K = 3 & 18.30M & 0.54s & \textbf{0.763} & \textbf{0.350} & \textbf{0.068}\\
         K = 4 & 18.30M & 0.66s & 0.761 & 0.356 & \textbf{0.068}\\
         K = 5 & 18.30M & 0.90s & 0.755 & 0.368 & 0.070\\
         \bottomrule
    \end{tabular}
    \label{tab:stage}
\end{table}

\begin{table}[t]
    \centering
    \setlength\tabcolsep{3pt}
    \caption{Ablation studies on main components. Exp.1 denotes using entire semantics. Exp.2 denotes generating motion with given frames only. Exp.3 denotes replacing Text-Frame Guided Block with the vanilla transformer decoder. Exp.4/5 denotes replacing Fusion Module with addition/concatenation fusion.}
    \begin{tabular}{cccccc}
        \bottomrule
         \multirow{2}{*}{Exp.} & &   \multicolumn{2}{c}{MotionCLIP} & \multicolumn{1}{c}{\multirow{2}{*}{AVE$^{Root}\downarrow$}} \\ \cline{3-4}
           &  & Top-1$\uparrow$ & \multicolumn{1}{c}{FID$^\star\downarrow$} & \multicolumn{1}{c}{}   \\ \toprule
         1 & Entire Semantics & 0.776 & 2.019 & 0.179 \\ 
         2 & Given frames to Motion & 0.528 & 1.601 & 0.125 \\
         3 & Vanilla Transformer Decoder & \textbf{0.793} & 1.112 & 0.082 \\
         4 & Addition Fusion & 0.762 & 0.590 & 0.075 \\
         5 & Concatenation Fusion & 0.761 & 0.520 & 0.078 \\ \hline
          & Our full PMG & 0.763 & \textbf{0.350} & \textbf{0.068} \\
         \toprule
    \end{tabular}
    \label{tab:components}
\end{table}

\begin{table}[t]
    \centering
    \caption{Effect of the number of given frames on HumanML3D\cite{t2m}. The best result is in blood, and the second best is in underline.}
    \setlength\tabcolsep{3pt}
    \begin{tabular}{ccccccc}
        \toprule[1pt]
        \multirow{2}{*}{\makecell{Number of \\ given frames}} & \multicolumn{2}{c}{MotionCLIP} & \multicolumn{2}{c}{T2M-Evaluator\cite{t2m}} & \multicolumn{1}{c}{\multirow{2}{*}{AVE$^{Root}\downarrow$}}  \\ \cmidrule(lr){2-3} \cmidrule(lr){4-5}
         &  \multicolumn{1}{c}{R-Top1$\uparrow$} & \multicolumn{1}{c}{FID$^\star\downarrow$} & \multicolumn{1}{c}{R-Top1$\uparrow$} & \multicolumn{1}{c}{FID$\downarrow$}  \\ \midrule
         0 & 0.720 & 1.569 & 0.521 & 0.061 & 0.126\\
         1 & 0.757 & 0.401 & 0.535 & 0.022 & 0.080 \\ 
         2 & 0.763 & 0.350 & \pmb{0.536} & 0.018 & 0.068 \\
         3 & \pmb{0.765} & 0.322 & 0.532 & 0.016 & 0.059 \\
         4 & 0.760 & \pmb{0.309} & 0.532 & \pmb{0.013} & \pmb{0.048}\\
         \bottomrule[1pt]
    \end{tabular}
    \label{tab:num_frame}
\end{table}

\begin{figure}[t]
    \centering
    \includegraphics[width=0.9\linewidth]{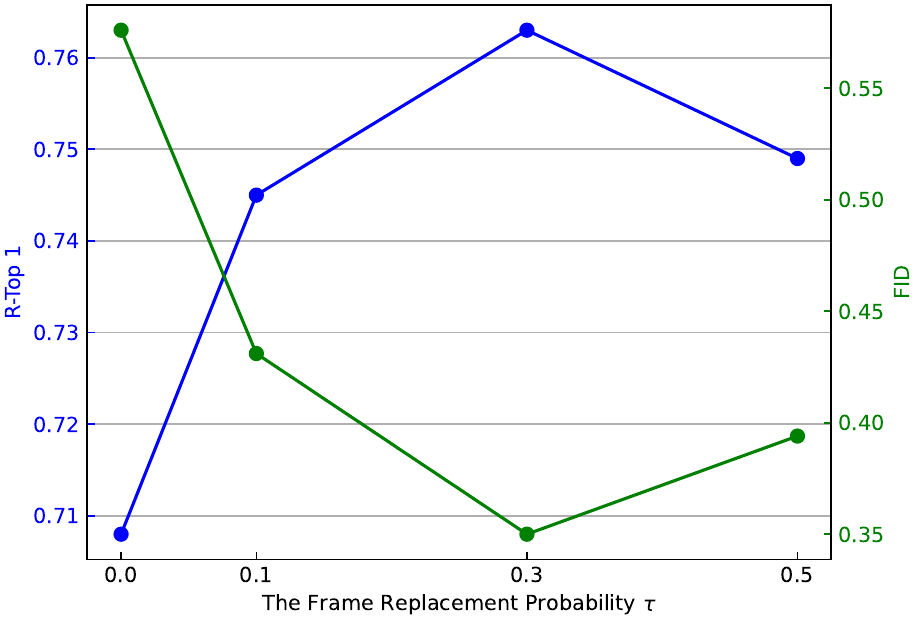}
    \caption{Effect of the Frame Replacement Probability $\tau$ on the HumanML3D Dataset. To alleviate the training-testing gap, we randomly replace training motions with those generated by the EMA model of our PMG, using a replacement probability $\tau$ during training. The displayed FID and R-Top 1 scores are evaluated by our MotionCLIP.}
    \label{fig:tau}
\end{figure}

\begin{table}[t]
    \centering
    \caption{Effect of text replacement probability $\eta$ on HumanML3D\cite{t2m}. The best result is in blood.}
    \setlength\tabcolsep{3pt}
    \begin{tabular}{ccccccc}
        \toprule[1pt]
        \multirow{2}{*}{} & \multicolumn{2}{c}{MotionCLIP} & \multicolumn{2}{c}{T2M-Evaluator\cite{t2m}} & \multicolumn{1}{c}{\multirow{2}{*}{AVE$^{Root}\downarrow$}}  \\ \cmidrule(lr){2-3} \cmidrule(lr){4-5}
         &  \multicolumn{1}{c}{R-Top1$\uparrow$} & \multicolumn{1}{c}{FID$^\star\downarrow$} & \multicolumn{1}{c}{R-Top1$\uparrow$} & \multicolumn{1}{c}{FID$\downarrow$}  \\ \midrule
         \rowcolor{orange!30} $\eta=0.1$ & \pmb{0.763} & \pmb{0.350} & \pmb{0.536} & \pmb{0.018} & \pmb{0.068}\\
         $\eta=0.2$ & 0.761 & 0.364 & 0.533 & \pmb{0.018} & 0.069 \\ 
         $\eta=0.3$ & 0.755 & 0.411 & 0.533 & 0.021 & 0.072 \\
         $\eta=0.4$ & 0.743 & 0.538 & 0.528 & 0.040 & 0.075 \\
         $\eta=0.5$ & 0.739 & 0.689 & 0.527 & 0.047 & 0.089\\
         \bottomrule[1pt]
    \end{tabular}
    \vspace{-0.2cm}
    \label{tab:text}
\end{table}

\begin{table}[t]
    \centering
    \caption{Effect of the guidance scale $s$ on HumanML3D\cite{t2m}. The best result is in blood, and the second best is in underline.}
    \setlength\tabcolsep{3pt}
    \begin{tabular}{ccccccc}
        \toprule[1pt]
        \multirow{2}{*}{\makecell{Guidance\\scale}} & \multicolumn{2}{c}{MotionCLIP} & \multicolumn{2}{c}{T2M-Evaluator\cite{t2m}} & \multicolumn{1}{c}{\multirow{2}{*}{AVE$^{Root}\downarrow$}}  \\ \cmidrule(lr){2-3} \cmidrule(lr){4-5}
         &  \multicolumn{1}{c}{R-Top1$\uparrow$} & \multicolumn{1}{c}{FID$^\star\downarrow$} & \multicolumn{1}{c}{R-Top1$\uparrow$} & \multicolumn{1}{c}{FID$\downarrow$}  \\ \midrule
         2.0 & 0.753 & 0.391 & 0.527 & 0.046 & \textbf{0.061} \\ 
         2.5 & 0.760 & \pmb{0.341} & 0.531 & 0.027 & \underline{0.063} \\
         \rowcolor{orange!30} 3.0 & \underline{0.763} & \underline{0.350} & \pmb{0.536} & \pmb{0.018} & 0.068 \\
         3.5 & \pmb{0.764} & 0.400 & 0.534 & \underline{0.020} & 0.071 \\
         4.0 & \pmb{0.764} & 0.490 & \underline{0.535} & 0.028 & 0.075\\
         4.5 & \pmb{0.764} & 0.620 & 0.533 & 0.046 & 0.079\\
         5.0 & 0.760 & 0.763 & 0.530 & 0.072 & 0.083 \\
         \bottomrule[1pt]
    \end{tabular}
    \vspace{-0.2cm}
    \label{tab:guidance}
\end{table}

\vspace{0.2cm}
\noindent\textbf{Effect of Generation Stage $K$.}
We evaluate our PMG with different generation stages, and results are shown in \Cref{tab:stage}. $K = 1$ indicates there is no frame partition, and PMG generates all frames at once. Compared with $K = 1$, $K > 1$ achieves significant improvements on FID and AVE$^{Root}$, demonstrating the effectiveness of generating motion frames progressively based on the distance to the nearest given frame. We also evaluate the inference times with different generation stages. As shown in \Cref{tab:stage}, the inference time increases when $K$ increases. We claim that a small $K$ is sufficient to achieve good performance on the HumanML3D dataset, while a larger $K$ increases inference time and slightly degrades performance. Furthermore, our PMG is lighter than both MotionDiffuse\cite{md} and T2M-GPT \cite{t2m-gpt}.

\vspace{0.2cm}
\noindent\textbf{Effectiveness of Main Components.}
To validate the effectiveness of PMG's components, we conduct the following experiments: 1) Using entire semantic information for generation during each stage. 2) Generating motion with given frames guidance only. 3) Replacing our Text-Frame Guided Block with the vanilla transformer decoder, where both \textit{key} and \textit{value} are the concatenation of obtained frames and frame-aware semantics. 4) Replacing our Fusion Module with addition fusion. 5) Replacing our Fusion Module with concatenation fusion.

As shown in \Cref{tab:components}, Exp.1 achieves implausibly high results on R-Top1 but extremely poor results on FID, validating our conjecture that the generated motions are overfitting on the text but losing realism. Exp.2 achieves extremely poor results on R-Top1 and FID but not bad AVE$^{Root}$, probably because the model is unable to generate complex motion and thus only generates frames that are very similar to the given frames. Exp.2 shows that only using given frames is not able to generate the target motion. Exp.3, 4, and 5 demonstrate the effectiveness of our proposed Text-Frame Guided Block.

\vspace{0.2cm}
\noindent\textbf{Effect of the Number of Given Frames.}
To evaluate the effect of the number of given frames, we test our PMG with different numbers of input frames.  As shown in \Cref{tab:num_frame}, providing more frames leads to higher performance. \la{Importantly, thanks to our special training strategy, our method is capable of generating human motion even without any given frames. Therefore, when users employ our PMG, we can first generate an initial human motion for them to select and adjust. Then, the user can input several adjusted frames back into PMG to generate the final motion.} Note that the trend of results evaluated by T2M-Evaluator \cite{t2m} differs due to its limited ability to discriminate between generated and real motions.

\vspace{0.2cm}
\noindent\textbf{Ablation Studies on Pseudo-frames Replacement Strategy.} To investigate the influence of the training-testing discrepancy, we train our PMG with the frame replacement probability $\tau = 0$. We train our PMG with different $\tau$ to evaluate its influence. As shown in \Cref{fig:tau}, $\tau = 0$ achieves the worst results, demonstrating there is indeed a training-testing gap. Our PMG achieves the best performance when $\tau = 0.3$.

\vspace{0.2cm}
\noindent\textbf{Effect of Guidance Scale $s$.}
The guidance scale $s$ in Eq. 10 controls the strength of the condition. To investigate the effect of the guidance scale $s$, we evaluate our PMG with different values of $s$.  As shown in \Cref{tab:guidance}, our PMG performs best when $s = 3$, while a higher $s$ results in poorer performance.

\vspace{0.2cm}
\noindent\textbf{Effect of Text Replacement Probability $\eta$.}
Text replacement probability $\eta$ controls the ratio between $\bar{\mathbf{c}}^k$ and $\hat{\mathbf{c}}^k$. \Cref{tab:text} shows the performance with different text replacement probabilities. Results show that higher text replacement probabilities lead to poorer performance; thus, we select $\eta = 0.1$ as our default setting.

\hl{

\vspace{0.2cm}
\noindent\textbf{Whether the obtained semantics are indeed frame-aware?} We analyzed the attention weights of a sample to verify whether the obtained semantics are indeed frame-aware. Specifically, for the sample shown in \Cref{fig:attn}, we selected three direction-indicating words, \ie, ``right'', ``forward'', and ``left'', from the given text for analysis. Correspondingly, we segmented the motion into four parts, with each part representing a specific walking direction. We then quantified the average attention weights $s_p$ between each direction-indicating word (token) and the given frames (tokens) in each motion part.  \Cref{tab:attn} shows the average attention weights. We find that each direction-indicating word pays more attention to the corresponding motion part, demonstrating that the obtained semantics are indeed frame-aware. Please refer to the Appendix for more details.
}

\begin{figure*}[t]
    \centering
    \includegraphics[width=\linewidth]{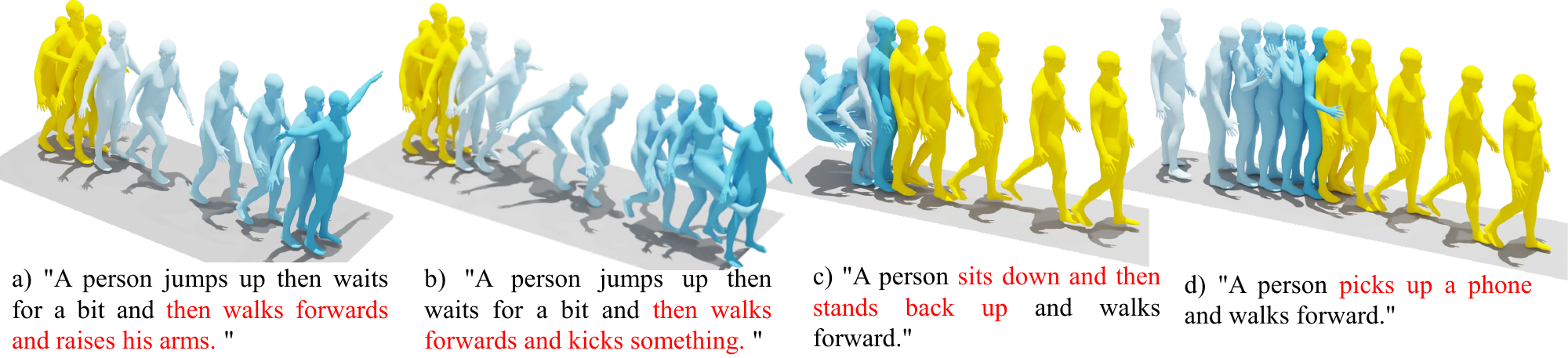}
    \caption{Our PMG can also perform motion inpainting at the frame level. When provided with some frames (indicated in yellow) and text prompts, our PMG can generate the remaining frames. The generated motions are indicated in chronological order, represented by a color gradient from gray to blue.}
    \label{fig:visual5}
\end{figure*}

\begin{figure}[]
    \centering
    \includegraphics[width=0.8\linewidth]{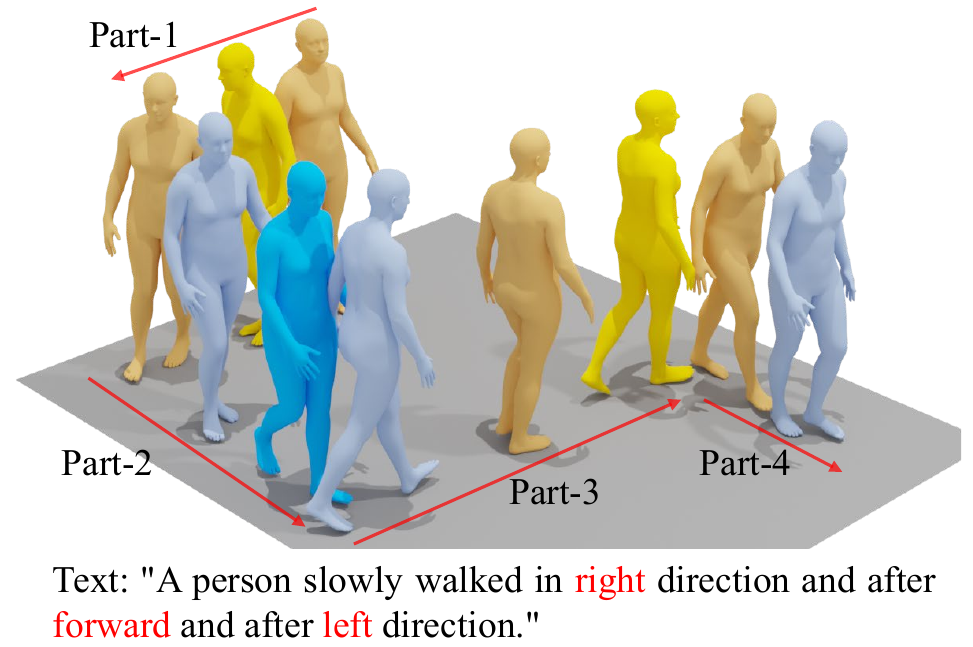}
    \caption{Visualization of the motion and the given text. The red lines indicate which frames belong to each motion part. The number of frames in each motion part are 80, 40, 52, and 24, respectively.}
    \label{fig:attn}
\end{figure}

\la{
\subsection{Application}
Due to our progressive motion generation paradigm and specialized training strategy, our PMG can be easily extended to motion temporal inpainting, as shown in \Cref{fig:visual5}. Specifically, by treating existing motion frames as given frames and providing text prompts for the content that needs to be modified, our model can generate motions consistent with the provided text. Moreover, the generated motions exhibit continuous and smooth transitions in their actions, seamlessly integrating with the existing motions.
}

\begin{table}[]
    \centering
    \caption{Average attention weights between the three direction-indicating words and the given frames (tokens) in each motion part. The highest attention weight is highlighted in bold. Note that the total number of frames is 196, so the uniform attention weight is 0.0051.}
    \begin{tabular}{ccccc}
      \toprule[1pt]
        & Part-1 & Part-2 & Part-3 & Part-4 \\
      \midrule
      ``right'' &  \textbf{0.0094} & 0.0053 & 0.0007 & 0\\
      ``forward'' &  0.0042 & \textbf{0.0120} & 0.0031 & 0.0001\\
      ``left'' & 0.0028 & 0.0042 & \textbf{0.0103} &  0.0031\\
      \toprule[1pt]
    \end{tabular}
    \label{tab:attn}
\end{table}


\section{Conclusion}
\hl{
In this work, we explore a new Text-Frame-to-Motion (TF2M) generation task that aims to achieve more controllable motion generation based on text and a few frames describing human postures. For this purpose, we propose a Progressive Motion Generation (PMG) method to progressively generate motion frames from those with low uncertainty to those with high uncertainty across multiple stages. 
Compared to the previous SOTA method, our method outperforms ReMoDiffuse$^\bullet$ with an FID of 0.101 (vs. 0.152) on the KIT-ML dataset and 0.018 (vs. 0.095) on the HumanML3D dataset when using 2 given frames.
Experiments show that our method can control the generated motion with only a few given frames, and that a stronger evaluator is urgently needed to assess the generated motions.

\noindent \textbf{Future Work.} Although our method is capable of generating realistic and coherent human motions, it may produce ambiguities in certain cases. Therefore, employing metaheuristic-based, nature-inspired optimization algorithms or reinforcement learning methods could enhance the motion generation process—particularly in managing frame transitions and resolving ambiguities in incomplete sequences. In the future, we plan to further improve the generation quality by building upon the insights from these works \cite{khan2021control, khan2022human, khan2022obstacle}.
}

\section{Acknowledgments}
This work was supported partially by NSFC(92470202, U21A20471), National Key Research and Development Program of China (2023YFA1008503), Guangdong NSF Project (No. 2023B1515040025).

\bibliographystyle{IEEEtran}
\bibliography{reference}

\clearpage

\appendices

\twocolumn[{
\begin{center}
  \Large\bfseries Appendix of ``Progressive Human Motion Generation \\Based on Text and Few Motion Frames'' 
  \vspace{1em} 
\end{center}
}] 

\section{Overview}
In the appendix, we provide additional details about both our method and experimental results.  In \Cref{sec:details of pmg}, we introduce in detail the generation scheme of our PMG.  In \Cref{sec:more visual}, we present more human motions generated by our PMG.  In \Cref{sec:details of motionclip}, we provide further details about our proposed MotionCLIP. In \Cref{sec: full results}, we present the complete results of our PMG and the compared methods evaluated using our MotionCLIP and T2M-Evaluator~\cite{t2m}. \hl{Finally, we provide the details of ablation study to demonstrate that the obtained semantics are frame-aware, and our user study in \Cref{sec:attn} and \Cref{sec:userstudy}.}

 \begin{figure*}[]
    \centering
    \includegraphics[width=\linewidth]{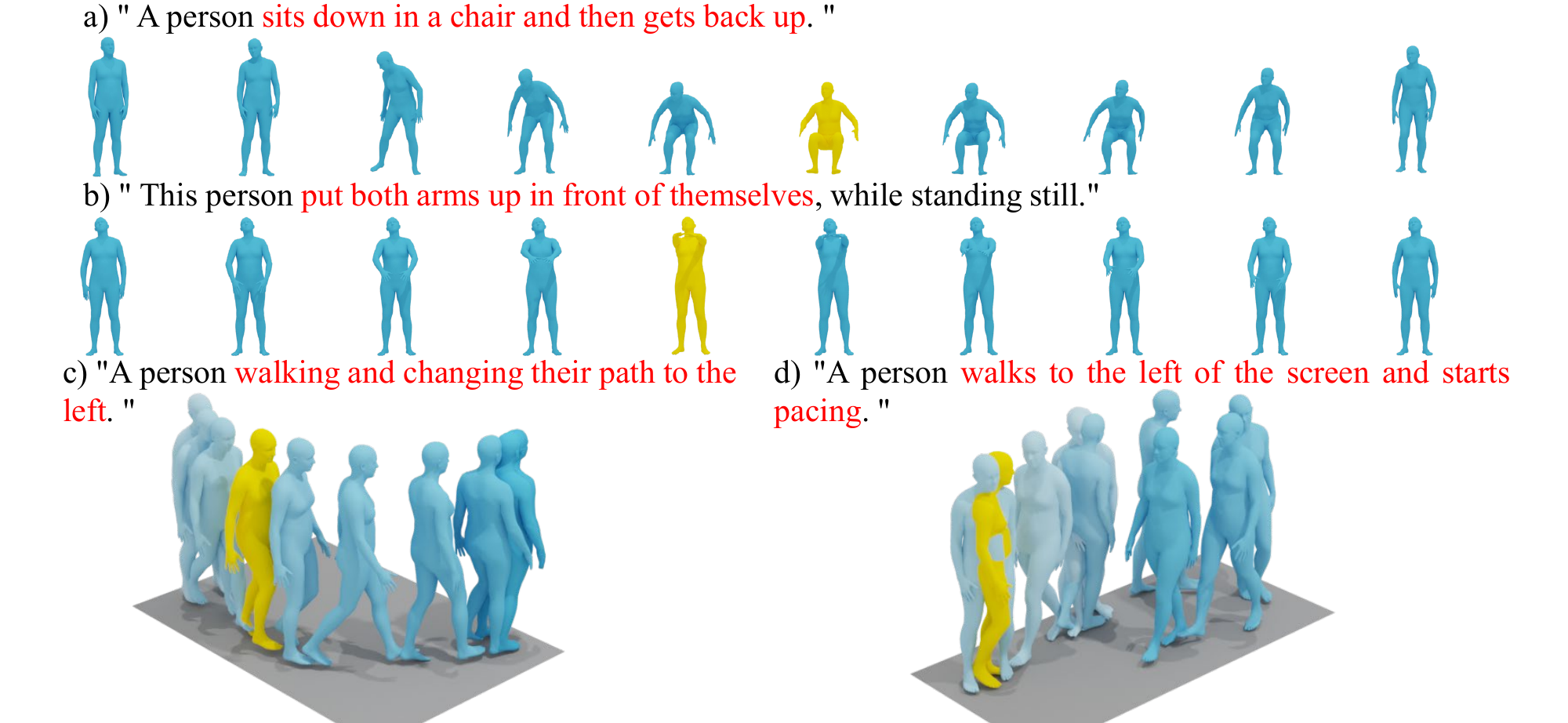}
    \caption{More human motions generated by our PMG.}
    \label{fig:visual4}
\end{figure*}

\section{Details of Progressive Motion Generation}\label{sec:details of pmg}

\noindent\textbf{Generation Scheme.}
\Cref{algo:sample} illustrates our motion generation scheme. We generate motion via UniPC \cite{zhao2023unipc} and $T_g$ is set as 10 during generation.

\begin{algorithm} 
    \caption{Motion Generation Scheme of PMG} 
    \renewcommand{\algorithmicrequire}{\textbf{Input:}} 
    \renewcommand{\algorithmicensure}{\textbf{Output:}} 
    \begin{algorithmic}[1] 
    {\small 
        \REQUIRE Given frames $\{f_i\}$ with their time positions $\{p_i\}$; text descriptions $\{w_i\}$; target motion length $N$; guidance $s$; number of stages $K$; diffusion steps $T_g$; pretrained PMG parameters $\theta$; 
        \STATE Divide the positions of target motion frames $P$ of length $N$ into $K$ groups using Eqs. 5 and 6; 
        \STATE $M^0 \leftarrow \{f_i\}$, $P^0 \leftarrow \{p_i\}$;
        \FOR{$k = 1, 2, ..., K$}  
            \STATE $\mathbf{x}_{T_g}^k \in \mathcal{R}^{|M^k| \times d_m} \sim \mathcal{N}(\textbf{0},\mathbf{I})$;
            \FOR{$t = T_g, T_{g-1}, ..., 1$}
                \STATE $\hat{\mathbf{c}}^k \leftarrow \{M^i\}_{i=0}^{k-1} \cup \{P^i\}_{i=0}^{k} \cup \varnothing$;
                \STATE $\bar{\mathbf{c}}^k \leftarrow \{M^i\}_{i=0}^{k-1} \cup \{P^i\}_{i=0}^{k} \cup  \{w_i\}$;
                \STATE Calculate the noise $\hat{\epsilon}_\theta(\mathbf{x}^k_t, t, \mathbf{c}^k)$ using Eq. 10; 
                \STATE Update $\mathbf{x}^k_{t-1}$ using Eqs. 3 and 4; 
            \ENDFOR 
        \STATE $M^k \leftarrow \mathbf{x}^k_0$; 
        \ENDFOR \STATE Obtain the final motion $M \leftarrow M^0 \cup ... \cup M^{K}$ by combining all stages; 
    } 
    \label{algo:sample}
    \end{algorithmic} 
\end{algorithm}

\section{More Visual Examples.} \label{sec:more visual}
We provide more human motions generated by our PMG, as shown in \Cref{fig:visual4}.

\section{Details of MotionCLIP} \label{sec:details of motionclip}
To more accurately evaluate the quality of generated motions, we designed a model called MotionCLIP, inspired by CLIP \cite{clip}, which can map matched text-motion pairs into closely aligned feature vectors in latent space. Our MotionCLIP consists of a motion encoder and a text encoder. The motion encoder is implemented using $L_3$ transformer encoders, while the text encoder utilizes a pretrained language model. During training, the matched text-motion pairs serve as positive samples, and randomly constructed mismatched text-motion pairs act as negative samples. We then train MotionCLIP using the contrastive loss from CLIP \cite{clip}. Thus, MotionCLIP learns to pull together the features of matched pairs and push apart the features of mismatched pairs. Additionally, we provide a figure in the supplementary material to illustrate our MotionCLIP.

Figure 5 in our main manuscript illustrates the architecture of our MotionCLIP. $L_3$ is set to 8, and we use the pretrained RoBERTa \cite{roberta} as our language model. We train our MotionCLIP with a batch size of 256 for 30 epochs on a single RTX 3090 GPU. The Adam optimizer is used with a learning rate of $10^{-4}$. We train our MotionCLIP in two stages. During the first 10 epochs, we freeze the pretrained language model and train the other parameters. Then, we unfreeze the pretrained language model and train all parameters together.

\section{Full Quantitative Results} \label{sec: full results}
The full results evaluated by our MotionCLIP are shown in \Cref{tab:hml3d-mc} and \Cref{tab:kit-mc}. As shown in both tables, our PMG significantly outperforms existing works when only one frame is given, especially in FID and R-Top1. Additionally, the full results evaluated by T2M-Evaluator \cite{t2m} are shown in \Cref{tab:hml3d-t2m} and \Cref{tab:kit-t2m}. Due to the limited discriminatory power of T2M-Evaluator between generated and authentic motions, the R-Top1 performance gap between our PMG and existing methods is smaller when assessed by T2M-Evaluator \cite{t2m} compared to our MotionCLIP. However, the performances on FID, MoBert \cite{best-metric} alignment score, and AVE$^{Root}$ still demonstrate the effectiveness of our PMG. On the other hand, since the diversity of generated motions when given the same frames and text (evaluated by the Multimodality metric, MM.) is constrained by the given frames, the MM of our PMG is slightly lower than that of other methods. Moreover, providing more given frames leads to lower MM.

\hl{
\section{Details of the Ablation Study} \label{sec:attn}
Since it is challenging to directly analyze the extent to which frame-aware semantics are related to the given frames, we analyzed the attention weights of a sample to verify whether the obtained semantics are indeed frame-aware. 

\noindent \textbf{Why use the attention weights to verify whether the obtained semantics are indeed frame-aware?} The attention weight between a semantic (text) token and a given frame token can be regarded as a metric to gauge how much information the semantic token extracts from the given frame token. Therefore, we provide an intuitive example of attention weights to demonstrate that the obtained semantics are frame-aware.

\noindent  \textbf{How to calculate the attention weights.} We computed the attention weights between key words and the given frames from the last cross-attention layer of the Frame-aware Semantics Decoder. Specifically, for the sample shown in Figure 10 of our manuscript, we selected three direction-indicating words, \ie, ``right'', ``forward'', and ``left'', from the given text for analysis. Correspondingly, we segmented the motion into four parts, with each part representing a specific walking direction. We then quantified the average attention weights $s_p$ between each direction-indicating word (token) and the given frames (tokens) in each motion part using the following formula:
    \begin{equation}
        s_p = \frac{\sum_{k=l_p}^{k=r_p}{a_k}}{r_p-l_p+1},
    \end{equation}
    where the range $[l_p, r_p]$ indicates which frames belong to part-$p$, and $a_k$ represents the attention weight between the word and the $k$-th given frame.

\section{Details of the User Study}\label{sec:userstudy}
\noindent \textbf{How to obtain hard-to-generate samples.} Since we lack explicit labels for generation difficulty to distinguish between easily and hard-to-generate samples, we propose using the R-Top1 metric to assess the ease of generating a given text. Specifically, R-Precision calculates the motion-retrieval precision when given one motion and 32 texts (1 real text and 31 randomly sampled mismatched texts). Thus, R-Top1 indicates how closely the model-generated sample aligns with the given text. Consequently, when retrieving with a model-generated sample, if the given text does not appear within the Top1 candidates, we consider this a hard-to-generate sample.

\noindent\textbf{The details of user study.} We use text prompts and 2 annotated given frames from the HumanML3D-Sub dataset to feed four models (including our method, OmniControl$^\bullet$, ReMoDiffuse$^\bullet$, and BAMM$^\bullet$), and then identify the hard-to-generate samples for all  models (using the method mentioned above). From these, we select 30 test texts. Each model then generates 3 motions for each test text. Thus, we obtain 90 generated samples for each model, and then perform 1-on-1 comparisons. In side-by-side comparisons of our method with existing methods, 25 participants are asked to select the one that better aligns with the given text or shows higher motion quality. 

\noindent \textbf{Analysis of the user study.} As shown in Figure 7, we observe that even though all samples are inherently challenging to generate (hard-to-generate samples), our method is still preferred by users most of the time. In direct comparisons against BAMM$^\bullet$, our method demonstrates advantages in both motion quality and text alignment, garnering user preference rates of 59.4\% and 56.9\% respectively. Furthermore, we notice that our model exhibits a relatively stronger performance in text alignment compared to motion quality.  This suggests that while our method excels at generating motions that are highly aligned with the given text prompts, the improvement in overall motion quality, though present, is not as pronounced.
}

\begin{table*}[ht]
    \centering
    \caption{Quantitative results evaluated by our MotionCLIP on HumanML3D\cite{t2m} test set. Following T2M\cite{t2m}, we repeat the evaluation \textbf{20} times and report the average with a $95\%$ confidence interval. We evaluate the released checkpoints of some works using new metrics, denoted by $*$. $\bullet$ denotes that this model is trained on the TF2M setting. The best result is in bold, the second best result is underlined. $\rightarrow$ indicates that values closer to those of real motion are better.}
    \vspace{-0.2cm}
    \setlength\tabcolsep{3pt}
    \resizebox{\linewidth}{!}{
    \begin{tabular}{ccccccccc}
        \toprule[1pt]
        \multirow{2}{*}{\makecell{\#Given\\frames}} & \multicolumn{1}{c}{\multirow{2}{*}{Methods}} & \multicolumn{7}{c}{MotionCLIP} \\ \cline{3-9}
         & \multicolumn{1}{c}{} &  R-Top1$\uparrow$ & R-Top2$\uparrow$ & R-Top3$\uparrow$ &  FID$^\star\downarrow$ & MM-Dist$\downarrow$ & \multicolumn{1}{c}{Diversity$\rightarrow$} & \multicolumn{1}{c}{MM. $\uparrow$}\\ \midrule
        - & \textbf{Real motion} & $0.752^{\pm.002}$ & $0.880^{\pm.002}$ & $0.924^{\pm.002}$ & $0.001^{\pm.000}$ & $0.908^{\pm.001}$ & $1.386^{\pm.004}$ & -\\ \hline 
        \multirow{5}{*}{0} 
        & MotionDiffuse\cite{md} & $0.665^{\pm.002}_*$ & $0.814^{\pm.002}_*$ & $0.874^{\pm.002}_*$ & $3.996^{\pm.040}_*$ & $1.362^{\pm.006}_*$ & $1.362^{\pm.005}$ & -\\
         & T2M-GPT\cite{t2m-gpt} & $0.661^{\pm.003}_*$ & $0.803^{\pm.003}_*$ & $0.860^{\pm.002}_*$ & $1.386^{\pm.023}_*$ & $0.954^{\pm.001}_*$ & $\underline{1.385}^{\pm.003}_*$ & $0.489^{\pm.028}_*$\\
         & ReMoDiffuse\cite{zhang2023remodiffuse} & $0.661^{\pm.003}_*$ & $0.806^{\pm.002}_*$ & $0.864^{\pm.002}_*$ & $1.981^{\pm.020}_*$ & $0.969^{\pm.001}_*$ & $\underline{1.387}^{\pm.004}_*$ & $\pmb{0.726}^{\pm.026}$\\ 
         & MoMask\cite{momask} & $0.722^{\pm.003}_*$ & $0.862^{\pm.002}_*$ & $0.911^{\pm.002}_*$ & $1.344^{\pm.021}_*$ & $0.930^{\pm.001}_*$ & $1.381^{\pm.005}_*$ & $0.380^{\pm.013}_*$\\
         & BAMM\cite{bamm} & $0.729^{\pm.004}_*$ & $0.865^{\pm.003}_*$ & $0.914^{\pm.002}_*$ & $1.369^{\pm.019}_*$ & $0.927^{\pm.001}_*$ & $1.382^{\pm.004}_*$ & $0.394^{\pm.026}_*$\\
         \hline
         & ReMoDiffuse$^\bullet$\cite{zhang2023remodiffuse} & $0.663^{\pm.002}$ & $0.811^{\pm.002}$ & $0.870^{\pm.002}$ & $1.923^{\pm.030}$ & $0.969^{\pm.001}$ & $1.377^{\pm.005}$ & $\underline{0.692}^{\pm.017}$\\
         & OmniControl$^\bullet$\cite{omnicontrol} & $0.598^{\pm.004}$ & $0.743^{\pm.003}$ & $0.798^{\pm.002}$ & $2.681^{\pm.051}$ & $1.216^{\pm.007}$ & $1.370^{\pm.006}$ & $0.493^{\pm.031}$ \\
         & BAMM$^\bullet$\cite{bamm} & $0.721^{\pm.004}$ & $0.861^{\pm.002}$ & $0.909^{\pm.001}$ & $1.376^{\pm.024}$ & $0.931^{\pm.001}$ & $1.379^{\pm.005}$ & $0.478^{\pm.023}$\\
         \rowcolor{orange!30} \cellcolor{white} \multirow{-3}{*}{1} &  Our PMG & $\underline{0.757}^{\pm.002}$ & $\underline{0.886}^{\pm.002}$ & $\underline{0.928}^{\pm.002}$ & $\underline{0.401}^{\pm.004}$ & $\underline{0.906}^{\pm.001}$ & $1.383^{\pm.003}$ & $0.464^{\pm.016}$\\ 
         \hline
         & ReMoDiffuse$^\bullet$\cite{zhang2023remodiffuse} &
         ${0.689}^{\pm.002}$ & ${0.831}^{\pm.002}$ & ${0.886}^{\pm.002}$ & ${1.582}^{\pm.020}$ & $0.953^{\pm.001}$ & ${1.380}^{\pm.004}$ & $0.541^{\pm.018}$\\
         & OmniControl$^\bullet$\cite{omnicontrol}  & $0.622^{\pm.002}$ & $0.765^{\pm.003}$ & $0.803^{\pm.002}$ & $2.135^{\pm.043}$  & $1.165^{\pm.005}$ & $1.367^{\pm.009}$ & $0.461^{\pm.021}$ \\
         & BAMM$^\bullet$\cite{bamm} & $0.737^{\pm.001}$ & $0.872^{\pm.002}$ & $0.918^{\pm.001}$ & $1.210^{\pm.018}$ & $0.923^{\pm.001}$ & $1.374^{\pm.006}$ & $0.447^{\pm.029}$\\
         \rowcolor{orange!30} \cellcolor{white} \multirow{-3}{*}{2} & Our PMG & $\pmb{0.763}^{\pm.003}$ & $\pmb{0.891}^{\pm.002}$ & $\pmb{0.932}^{\pm.001}$ & $\pmb{0.350}^{\pm.004}$ & $\pmb{0.902}^{\pm.001}$ & $\pmb{1.386}^{\pm.004}$ & $0.423^{\pm.012}$\\
         \bottomrule[1pt]
    \end{tabular}
    }
    \vspace{-0.2cm}
    \label{tab:hml3d-mc}
\end{table*}

\begin{table*}[ht]
    \centering
    \caption{
    Quantitative results evaluated by our MotionCLIP on KIT-ML\cite{kit} test set. Following T2M\cite{t2m}, we repeat the evaluation \textbf{20} times and report the average with a $95\%$ confidence interval. We evaluate the released checkpoints of some works using new metrics, denoted by $*$. $\bullet$ denotes that this model is trained on the TF2M setting. The best result is in bold, the second best result is underlined. $\rightarrow$ indicates that values closer to those of real motion are better.}
    \vspace{-0.2cm}
    \setlength\tabcolsep{3pt}
    \resizebox{\linewidth}{!}{
    \begin{tabular}{cccccccccc}
        \toprule[1pt]
        \multirow{2}{*}{\makecell{\#Given\\frames}} & \multicolumn{1}{c}{\multirow{2}{*}{Methods}} & \multicolumn{7}{c}{MotionCLIP} \\ \cline{3-9}
         & \multicolumn{1}{c}{} &  R-Top1$\uparrow$ & R-Top2$\uparrow$ & R-Top3$\uparrow$ &  FID$^\star\downarrow$ & MM-Dist$\downarrow$ & \multicolumn{1}{c}{Diversity$\rightarrow$} & \multicolumn{1}{c}{MM. $\uparrow$}\\ \midrule
        - & \textbf{Real motion} & $0.536^{\pm.006}$ & $0.738^{\pm.004}$ & $0.832^{\pm.004}$ &  $0.013^{\pm.005}$ & $0.844^{\pm.001}$ & $1.359^{\pm.012}$ & -\\ \hline 
        \multirow{4}{*}{0} & MotionDiffuse\cite{md} & $0.567^{\pm.004}_*$ & $0.747^{\pm.005}_*$ & $0.831^{\pm.003}_*$ & $7.281^{\pm.085}_*$ & $0.844^{\pm.002}_*$ & $1.305^{\pm.007}_*$& -\\
         & T2M-GPT\cite{t2m-gpt} & $0.537^{\pm.005}_*$ & $0.727^{\pm.006}_*$ & $0.815^{\pm.005}_*$ & $2.703^{\pm.055}_*$ & $0.848^{\pm.003}_*$ & $1.350^{\pm.006}_*$ & $\pmb{0.657}^{\pm.034}_*$\\
         & ReMoDiffuse\cite{zhang2023remodiffuse} & $0.550^{\pm.006}_*$ & $0.747^{\pm.006}_*$ & $0.838^{\pm.006}_*$ & $2.332^{\pm.031}_*$ & $0.851^{\pm.002}_*$ & ${1.364}^{\pm.006}_*$ & $\underline{0.463}^{\pm.014}_*$\\ 
         & MoMask\cite{momask} & $0.568^{\pm.006}_*$ & $0.770^{\pm.005}_*$ & $0.858^{\pm.005}_*$ & $2.699^{\pm.016}_*$  & $0.845^{\pm.002}_*$ & $1.370^{\pm.004}_*$ & $0.347^{\pm.010}_*$\\
         \hline
         & ReMoDiffuse$^\bullet$\cite{zhang2023remodiffuse} & $0.531^{\pm.007}$ & $0.729^{\pm.006}$ & $0.826^{\pm.005}$ & $2.491^{\pm.040}$ & $0.858^{\pm.001}$ & $1.354^{\pm.006}$ & $0.382^{\pm.013}$\\
         & OmniControl$^\bullet$\cite{omnicontrol} & $0.541^{\pm.005}$ &  $0.744^{\pm.005}$ &  $0.819^{\pm.004}$ &  $2.795^{\pm.072}$ &  $0.847^{\pm.003}$ &  $1.351^{\pm.006}$ &  $0.396^{\pm.017}$\\
         \rowcolor{orange!30} \cellcolor{white} \multirow{-3}{*}{1} &  Our PMG & $\pmb{0.612}^{\pm.007}$ & $\pmb{0.805}^{\pm.006}$ & $\pmb{0.888}^{\pm.003}$ & $\underline{1.258}^{\pm.012}$ & $\underline{0.793}^{\pm.002}$ & $\underline{1.363}^{\pm.013}$ & $0.245^{\pm.008}$\\ 
         \hline
         & ReMoDiffuse$^\bullet$\cite{zhang2023remodiffuse} &
         ${0.562}^{\pm.006}$ & ${0.751}^{\pm.006}$ & ${0.843}^{\pm.006}$ & ${2.260}^{\pm.026}$ & $0.838^{\pm.002}$ & ${1.364}^{\pm.006}$ & $0.345^{\pm.007}$\\
         & OmniControl$^\bullet$\cite{omnicontrol} & $0.549^{\pm.004}$ &  $0.746^{\pm.003}$ &  $0.826^{\pm.006}$ &  $2.651^{\pm.041}$ &  $0.845^{\pm.003}$ &  $1.352^{\pm.008}$ &  $0.384^{\pm.013}$\\
         \rowcolor{orange!30} \cellcolor{white} \multirow{-3}{*}{2} & Our PMG & $\underline{0.603}^{\pm.007}$ & $\underline{0.798}^{\pm.006}$ & $\underline{0.884}^{\pm.003}$ & $\pmb{1.119}^{\pm.012}$ & $\pmb{0.790}^{\pm.001}$ & $\pmb{1.356}^{\pm.009}$ & $0.222^{\pm.006}$\\
         \bottomrule[1pt]
    \end{tabular}
    }
    \vspace{-0.2cm}
    \label{tab:kit-mc}
\end{table*}

\begin{table}[htbp]
  \centering
  \begin{minipage}[t]{1\textwidth}
    \centering
    \caption{Quantitative results evaluated by T2M-Evaluator\cite{t2m}, MoBERT\cite{best-metric}, and coordinate error metrics on HumanML3D\cite{t2m} test set. Following T2M\cite{t2m}, we repeat the evaluation \textbf{20} times and report the average with a $95\%$ confidence interval. We evaluate the released checkpoints of some works using new metrics, denoted by $*$. $\bullet$ denotes that this model is trained on the TF2M setting. The best result is in bold, the second best result is underlined. $\rightarrow$ indicates that values closer to those of real motion are better.}
    \vspace{-0.2cm}
    \setlength\tabcolsep{3pt}
    \resizebox{\linewidth}{!}{
    \begin{tabular}{ccccccccccc}
        \toprule[1pt]
        \multirow{2}{*}{\makecell{\#Given\\frames}} & \multicolumn{1}{c}{\multirow{2}{*}{Methods}} &  \multicolumn{1}{c}{\multirow{2}{*}{\makecell{MoBert \cite{best-metric} \\ Alignment$\uparrow$}}} & \multicolumn{1}{c}{\multirow{2}{*}{AVE$^{Root}\downarrow$}}  & \multicolumn{7}{c}{T2M Evaluator\cite{t2m}} \\ \cline{5-11}
         & \multicolumn{1}{c}{} & \multicolumn{1}{c}{} & \multicolumn{1}{c}{} &  R-Top1$\uparrow$ & R-Top2$\uparrow$ & R-Top3$\uparrow$ &  FID$\downarrow$ & MM-Dist$\downarrow$ & \multicolumn{1}{c}{Diversity$\rightarrow$}  & \multicolumn{1}{c}{MM.$\uparrow$}\\ \midrule
        - & \textbf{Real motion} & $0.621^{\pm.002}$ & 0 & $0.511^{\pm.003}$ & $0.703^{\pm.003}$ & $0.797^{\pm.002}$ & $0.002^{\pm.000}$ & $2.974^{\pm.008}$ & $9.503^{\pm.065}$ & - \\ \hline 
        \multirow{12}{*}{0}
         & T2M\cite{t2m} & - & - & $0.455^{\pm.003}$ & $0.636^{\pm.003}$ & $0.736^{\pm.002}$ & $1.087^{\pm.021}$ & $3.347^{\pm.008}$ & $9.175^{\pm.083}$ & $2.090^{\pm.083}$\\
         & MDM\cite{mdm} & - & - & $0.320^{\pm.005}$ & $0.498^{\pm0.004}$ & $0.611^{\pm.007}$ & $0.544^{\pm.044}$ & $5.566^{\pm.027}$ & $\underline{9.559}^{\pm.086}$ & $\underline{2.799}^{\pm.072}$\\
         & MLD\cite{mld} & - & - & $0.481^{\pm.003}$ & $0.673^{\pm.003}$ & $0.772^{\pm.002}$ & $0.473^{\pm.013}$ & $3.196^{\pm.010}$ & $9.724^{\pm.082}$ & $2.192^{\pm.079}$\\
         & MotionDiffuse\cite{md} & $0.561^{\pm.003}_*$ & $0.197^{\pm.043}_*$ & $0.491^{\pm.001}$ & $0.681^{\pm.001}$ & $0.782^{\pm.001}$ & $0.630^{\pm.001}$ & $3.113^{\pm.001}$ & $9.410^{\pm.049}$ & $0.730^{\pm.013}$\\
         & T2M-GPT\cite{t2m-gpt} & $0.597^{\pm.003}_*$ & $0.179^{\pm.032}_*$ & $0.491^{\pm.003}$ & $0.680^{\pm.003}$ & $0.775^{\pm.002}$ & $0.116^{\pm.004}$ & $3.118^{\pm.011}$ & $9.761^{\pm.081}$ & $1.856^{\pm.011}$\\
         & M2DM \cite{kong2023priority} & - & - & $0.497^{\pm.003}$ & $0.682^{\pm.002}$ & $0.763^{\pm.003}$ & $0.352^{\pm.005}$ & $3.134^{\pm.010}$ & $9.926^{\pm.073}$ & $\pmb{3.587}^{\pm.072}$\\
         & Fg-T2M\cite{fgt2m} & - & - & $0.492^{\pm.002}$ & $0.683^{\pm.003}$ & $0.783^{\pm.002}$ & $0.243^{\pm.019}$ & $3.109^{\pm.007}$ & $9.278^{\pm.072}$ & $1.164^{\pm.049}$\\
         & Att-T2M\cite{attt2m} & - & - & $0.499^{\pm.003}$ & $0.690^{\pm.002}$ & $0.786^{\pm.002}$ & $0.112^{\pm.006}$ & $3.038^{\pm.007}$ & $9.700^{\pm.090}$ & $2.452^{\pm.051}$\\ 
         & MotionGPT\cite{jiang2023motiongpt} & - & - & $0.492^{\pm.003}$ & $0.681^{\pm.003}$ & $0.778^{\pm.002}$ & $0.232^{\pm.008}$ & $3.096^{\pm.008}$ & $\pmb{9.528}^{\pm.071}$ & $2.008^{\pm.084}$\\
         & ReMoDiffuse\cite{zhang2023remodiffuse} & $0.589^{\pm.002}_*$ & $0.157^{\pm.026}_*$ & $0.510^{\pm.005}$ & $0.698^{\pm.006}$ & $0.795^{\pm.004}$ & $0.103^{\pm.004}$ & $2.974^{\pm.016}$ & $9.018^{\pm.075}$ & $1.795^{\pm.043}$\\
         & MoMask\cite{momask} & $0.621_*^{\pm.002}$ & $0.122_*^{\pm.064}$ &  ${0.521}^{\pm.002}$ & ${0.713}^{\pm.002}$ & ${0.807}^{\pm.002}$ & $0.045^{\pm.002}$ & ${2.958}^{\pm.008}$ & $9.685^{\pm.087}_*$ & $1.241^{\pm.040}$                 \\
         & BAMM\cite{bamm} & $0.625_*^{\pm.005}$ & $0.124_*^{\pm.047}$ &  ${0.522}^{\pm.003}$ & ${0.715}^{\pm.003}$ & ${0.808}^{\pm.003}$ & $0.055^{\pm.002}$ & ${2.936}^{\pm.077}$ & $9.636^{\pm.009}$ & $1.732^{\pm.055}$                 \\
         \hline
         & ReMoDiffuse$^\bullet$\cite{zhang2023remodiffuse} & $0.593^{\pm.003}$ & $0.158^{\pm.031}$ & $0.508^{\pm.003}$ & $0.688^{\pm.003}$ & $0.786^{\pm.002}$ & $0.105^{\pm.002}$ & $3.035^{\pm.012}$ & $9.248^{\pm.010}$ & $2.490^{\pm.127}$\\
         & OmniControl$^\bullet$\cite{omnicontrol} &  $0.565^{\pm.007}$ &  $0.195^{\pm.037}$ & $0.434^{\pm.005}$ &  $0.625^{\pm.004}$ &  $0.731^{\pm.002}$ &  $0.461^{\pm.032}$ &  $3.186^{\pm.018}$ &  $9.376^{\pm.046}$ &  $1.967^{\pm.046}$\\
         & BAMM$^\bullet$\cite{bamm} &  $0.624^{\pm.003}$ &  $0.120^{\pm.035}$ & $0.519^{\pm.004}$ &  $0.709^{\pm.003}$ &  $0.799^{\pm.002}$ &  $0.057^{\pm.003}$ &  $2.947^{\pm.061}$ &  $9.621^{\pm.076}$ &  $1.729^{\pm.051}$\\
         \rowcolor{orange!30} \cellcolor{white} \multirow{-4}{*}{1} & Our PMG & $\underline{0.646}^{\pm.002}$ & $\underline{0.080}^{\pm.001}$ & $\underline{0.535}^{\pm.003}$ & $\underline{0.730}^{\pm.002}$ & $\pmb{0.822}^{\pm.002}$ & $\underline{0.022}^{\pm.001}$ & $\underline{2.834}^{\pm.005}$ & $9.560^{\pm.092}$ & $1.673^{\pm.069}$\\ \hline
         & ReMoDiffuse$^\bullet$\cite{zhang2023remodiffuse} & $0.598^{\pm.003}$ & $0.150^{\pm.033}$  & $0.511^{\pm.005}$ & $0.698 ^{\pm.003}$ & $0.796^{\pm.002}$ & $0.095^{\pm.005}$ & $2.964^{\pm.008}$ & $9.197^{\pm.054}$ & $2.367^{\pm.010}$\\
         & OmniControl$^\bullet$\cite{omnicontrol} &$0.573^{\pm.005}$ & $0.186^{\pm.039}$ & $0.467^{\pm.003}$ & $0.667^{\pm.003}$ & $0.745^{\pm.004}$ & $0.265^{\pm.012}$ & $3.115^{\pm.019}$ & $9.247^{\pm.076}$ & $1.740^{\pm.038}$\\
         & BAMM$^\bullet$\cite{bamm} &  $0.631^{\pm.002}$ &  $0.102^{\pm.034}$ & $0.526^{\pm.002}$ &  $0.719^{\pm.002}$ &  $0.813^{\pm.002}$ &  $0.043^{\pm.002}$ &  $2.924^{\pm.039}$ &  $9.592^{\pm.063}$ &  $1.684^{\pm.047}$\\
         \rowcolor{orange!30} \cellcolor{white} \multirow{-4}{*}{2} & Our PMG & $\pmb{0.650}^{\pm.002}$ & $\pmb{0.068}^{\pm.001}$ & $\pmb{0.536}^{\pm.002}$ & $\pmb{0.731}^{\pm.003}$ & $\underline{0.821}^{\pm.002}$ & $\pmb{0.018}^{\pm.001}$ & $\pmb{2.832}^{\pm.009}$ & $\pmb{9.528}^{\pm.086}$ & $1.527^{\pm.051}$\\
         \bottomrule[1pt]
    \end{tabular}
    }
    \vspace{-0.2cm}
    \label{tab:hml3d-t2m}
  \end{minipage}
  \hfill
  \vspace{0.5cm}
  \begin{minipage}[t]{1\textwidth}
    \centering
    \caption{
    Quantitative results evaluated by T2M-Evaluator\cite{t2m}, MoBERT\cite{best-metric}, and coordinate error metrics on KIT-ML\cite{kit} test set. Following T2M\cite{t2m}, we repeat the evaluation \textbf{20} times and report the average with a $95\%$ confidence interval. We evaluate the released checkpoints of some works using new metrics, denoted by $*$. $\bullet$ denotes that this model is trained on the TF2M setting. The best result is in bold, the second best result is underlined. $\rightarrow$ indicates that values closer to those of real motion are better.
    } 
    \vspace{-0.2cm}
    \setlength\tabcolsep{3pt}
    \resizebox{\linewidth}{!}{
    \begin{tabular}{cccccccccc}
        \toprule[1pt]
        \multirow{2}{*}{\makecell{\#Given\\frames}} & \multicolumn{1}{c}{\multirow{2}{*}{Methods}} & \multicolumn{1}{c}{\multirow{2}{*}{AVE$^{Root}\downarrow$}}  & \multicolumn{7}{c}{T2M Evaluator\cite{t2m}} \\ \cline{4-10}
         & \multicolumn{1}{c}{} & \multicolumn{1}{c}{} &  R-Top1$\uparrow$ & R-Top2$\uparrow$ & R-Top3$\uparrow$ &  FID$\downarrow$ & MM-Dist$\downarrow$ & \multicolumn{1}{c}{Diversity$\rightarrow$}  & \multicolumn{1}{c}{MM.$\uparrow$}\\ \midrule
        - & \textbf{Real motion} & 0 & $0.424^{\pm.005}$ & $0.649^{\pm.006}$ & $0.779^{\pm.006}$ & $0.031^{\pm.004}$ & $2.788^{\pm.012}$ & $11.08^{\pm.097}$ & - \\ \hline 
         \multirow{11}{*}{0} & T2M\cite{t2m} & - & $0.361^{\pm.006}$ & $0.559^{\pm.007}$ & $0.681^{\pm.007}$ & $3.022^{\pm.107}$ & $3.488^{\pm.028}$ & $10.72^{\pm.145}$ & $1.482^{\pm.065}$\\
         & MDM\cite{mdm} & - & $0.164^{\pm.005}$ & $0.291^{\pm.004}$ & $0.396^{\pm.004}$ & $0.497^{\pm.021}$ & $9.191^{\pm.022}$ & $10.85^{\pm.109}$ & $1.907^{\pm.214}$\\
         & MLD\cite{mld} & - & $0.390^{\pm.008}$ & $0.609^{\pm.008}$ & $0.734^{\pm.007}$ & $0.404^{\pm.027}$ & $3.204^{\pm.027}$ & $10.80^{\pm.117}$ & $2.192^{\pm.071}$\\
         & MotionDiffuse\cite{md} & $0.496^{\pm.101}_*$ & $0.417^{\pm.004}$ & $0.621^{\pm.004}$ & $0.739^{\pm.004}$ & $1.954^{\pm.064}$ & $2.958^{\pm.005}$ & $\underline{11.10}^{\pm.143}$ & $0.730^{\pm.013}$\\
         & T2M-GPT\cite{t2m-gpt} & $0.443^{\pm.060}_*$ & $0.416^{\pm.006}$ & $0.627^{\pm.006}$ & $0.745^{\pm.006}$ & $0.514^{\pm.029}$ & $3.007^{\pm.023}$ & $10.92^{\pm.108}$ & $1.570^{\pm.039}$\\
         & M2DM \cite{kong2023priority} & - & $0.416^{\pm.004}$ & $0.628^{\pm.004}$ & $0.743^{\pm.004}$ & $0.515^{\pm.029}$ & $3.015^{\pm.017}$ & $11.42^{\pm.970}$ & $\pmb{3.325}^{\pm.370}$\\
         & Fg-T2M\cite{fgt2m} & - & $0.418^{\pm.005}$ & $0.626^{\pm.004}$ & $0.745^{\pm.004}$ & $0.571^{\pm.047}$ & $3.114^{\pm.015}$ & $10.93^{\pm.083}$ & $1.019^{\pm.029}$\\
         & Att-T2M\cite{attt2m} & - & $0.413^{\pm.006}$ & $0.632^{\pm.006}$ & $0.751^{\pm.006}$ & $0.870^{\pm.039}$ & $3.039^{\pm.021}$ & $10.96^{\pm.123}$ & $2.281^{\pm.047}$\\ 
         & MotionGPT\cite{jiang2023motiongpt} & - & $0.366^{\pm.005}$ & $0.558^{\pm.004}$ & $0.680^{\pm.006}$ & $0.510^{\pm.016}$ & $3.527^{\pm.021}$ & $10.35^{\pm.084}$ & $\underline{2.328}^{\pm.117}$\\
         & ReMoDiffuse\cite{zhang2023remodiffuse} &  $0.453^{\pm.100}_*$ & $0.427^{\pm.014}$ & $0.641^{\pm.004}$ & $0.765^{\pm.055}$ & $0.155^{\pm.006}$ & $2.814^{\pm.012}$ & $10.80^{\pm.105}$ & $1.239^{\pm.028}$\\
         & MoMask\cite{momask} & ${0.439}^{\pm.057}_*$ & ${0.433}^{\pm.007}$ & ${0.656}^{\pm.005}$ & $\underline{0.781}^{\pm.005}$ & $0.204^{\pm.011}$ & ${2.779}^{\pm.022}$ & $10.82^{\pm.091}_*$ &  $1.131^{\pm.043}$\\
         
         \hline
         & ReMoDiffuse$^\bullet$\cite{zhang2023remodiffuse} & $0.451^{\pm.069}$ & $0.426^{\pm.005}$ & $0.644^{\pm.006}$ & $0.764^{\pm.004}$ & $0.169^{\pm.010}$ & $2.860^{\pm.020}$ & $10.75^{\pm.086}$ & $1.682^{\pm.110}$\\
         & OmniControl$^\bullet$\cite{omnicontrol} & $0.468^{\pm.058}$ & $0.387^{\pm.007}$ & $0.572^{\pm.005}$  & $0.695^{\pm.007}$ & $0.632^{\pm.021}$ &  $3.115^{\pm.019}$ & $10.84^{\pm.051}$ & $1.729^{\pm.063}$\\
         \rowcolor{orange!30} \cellcolor{white} \multirow{-2}{*}{1} & Our PMG & $\underline{0.323}^{\pm.003}$ & $\pmb{0.453}^{\pm.005}$ & $\pmb{0.682}^{\pm.006}$ & $\pmb{0.800}^{\pm.006}$ & $\underline{0.118}^{\pm.004}$ & $\underline{2.637}^{\pm.015}$ & ${11.11}^{\pm.217}$ & ${1.030}^{\pm.045}$\\ \hline
         
         & ReMoDiffuse$^\bullet$\cite{zhang2023remodiffuse} & $0.431^{\pm.081}$  & $0.429^{\pm.006}$ & $0.656^{\pm.007}$ & $0.778^{\pm.003}$ & $0.152^{\pm.001}$ & $2.771^{\pm.016}$ & $10.74^{\pm.080}$ & $1.520^{\pm.093}$\\
         & OmniControl$^\bullet$\cite{omnicontrol} &   $0.462^{\pm.056}$ & $0.400^{\pm.006}$ & $0.610^{\pm.008}$ & $0.743^{\pm.007}$ & $0.565^{\pm.026}$ & $3.085^{\pm.025}$ & $10.67^{\pm.072}$ &   $1.560^{\pm.040}$\\
         \rowcolor{orange!30} \cellcolor{white} \multirow{-2}{*}{2} & Our PMG & $\pmb{0.308}^{\pm.003}$ & $\underline{0.451}^{\pm.007}$ & $\underline{0.678}^{\pm.007}$ & $\pmb{0.800}^{\pm.005}$ & $\pmb{0.101}^{\pm.002}$ & $\pmb{2.630}^{\pm.016}$ & $\pmb{11.09}^{\pm.127}$ & ${0.917}^{\pm.039}$\\
         \bottomrule[1pt]
    \end{tabular}
    }
    \vspace{-0.2cm}
    \label{tab:kit-t2m}
  \end{minipage}
\end{table}

\end{document}